%% file: main.tex
\documentclass[10pt]{article}
\usepackage[a4paper, total={6in, 8in}]{geometry}

\usepackage[comma,authoryear,round]{natbib}
\usepackage[hidelinks,
colorlinks=true,
linkcolor=blue,
citecolor=blue,
backref=page,
]{hyperref}
 \renewcommand*{\backrefalt}[4]{%
    \ifcase #1%
     \or (cited on page:~#2)%
     \else (cited on pages:~#2)%
    \fi%
    }

\usepackage{stmaryrd}
\usepackage{microtype}
\usepackage{subcaption}
\usepackage{graphicx}
\usepackage{booktabs} 
\usepackage{hyperref}
\usepackage{amsmath}
\usepackage{amssymb}
\usepackage{mathtools}
\usepackage{amsthm}

\usepackage{tikz}
\usetikzlibrary{positioning}

\theoremstyle{plain}
\newtheorem{theorem}{Theorem}[section]
\newtheorem{theorem*}{Theorem}[section]

\newtheorem{lemma}[theorem]{Lemma}
\newtheorem{lemma*}[theorem]{Lemma}

\theoremstyle{definition}

\theoremstyle{remark}

\newcommand{\cause}[1]{&{\color{gray}\downarrow{}\;{\small \text{#1}}} \nonumber\\}
\newcommand{\Pb}{\mathbf{P}}
\newcommand{\E}{\mathbf{E}}

\newcommand{\Var}{\mathbf{V}}
\newcommand{\Cov}{\mathbf{Cov}}

\newcommand{\R}{\mathbb{R}}
\newcommand{\I}{\mathbf{I}}
\newcommand{\V}{\mathbf{V}}

\newcommand{\N}{\mathcal{N}}
\newcommand{\cald}{\mathcal{D}}
\newcommand{\calm}{\mathcal{M}}

\newcommand{\calh}{\mathcal{H}}
\newcommand{\cale}{\mathcal{E}}
\newcommand{\calv}{\mathcal{V}}

\renewcommand{\S}{G}
\newcommand{\A}{A}
\newcommand{\D}{D}

\newcommand{\h}{h}

\newcommand{\hp}{\h^\text{priv}}

\newcommand{\fk}{\mathcal{F}_{k}}
\newcommand{\amargin}{\alpha}

\newcommand{\x}{x}
\newcommand{\X}{X}
\newcommand{\calx}{\mathcal{\X}}
\newcommand{\y}{y}
\newcommand{\Y}{Y}
\newcommand{\caly}{\mathcal{\Y}}
\newcommand{\s}{s}
\renewcommand{\S}{S}
\newcommand{\cals}{\mathcal{\S}}
\newcommand{\calf}{\mathcal{F}}
\newcommand{\call}{\mathcal{L}}
\newcommand{\yh}{\hat \y}

\newcommand{\z}{z}
\newcommand{\Z}{Z}

\newcommand{\acc}{\mathcal{A}}

\DeclareMathOperator*{\argmax}{arg\,max}
\DeclareMathOperator*{\argmin}{arg\,min}

\usepackage{authblk}

\title{On the Impact of Output Perturbation on Fairness\\ in Binary Linear Classification}
\author[1]{Vitalii Emelianov}
\author[1]{Micha\"el Perrot}
\affil[1]{Univ. Lille, Inria, CNRS, Centrale Lille,
UMR 9189 - CRIStAL, F-59000 Lille, France. 
}

\date{}

\begin{document}

\maketitle

\begin{abstract}
  \input{abstract.tex}
\end{abstract}

\section{Introduction}
\label{sec:intro}
\input{intro.tex}

\section{Setting and Notations}
\label{sec:model}
\input{model.tex}

\section{Individual Fairness of Private Models}
\label{sec:individual fairness}
\input{individual_fairness.tex}

\section{Disagreement of Non-Private and Private Models}
\label{sec:disagreement}
\input{disagreement.tex}

\section{Group Fairness of Private Models}
\label{sec:bound}
\input{group_fairness.tex}

\section{Using the Bounds in Various Settings}
\label{sec:auditing}
\input{various-settings.tex}

\section{Conclusion and Discussion}
\label{sec:conclusion}
\input{conclusion.tex}

\section*{Acknowledgments}
\label{sec:acknowledgments}
This work was supported by the Région Hauts de France (Projet STaRS \'Equité en apprentissage décentralisé respectueux de la vie privée) and by the French National Research Agency
(ANR) through the grant ANR-23-CE23-0011 (Project FaCTor). This work has benefited from French State aid managed by the Agence Nationale de la Recherche (ANR) under France 2030 program with the reference ANR-23-PEIA-005 (REDEEM project).

\bibliography{bibliography}

\newpage
\appendix

\section{Proofs}
\label{sec:proofs}
\input{./appendix/proofs.tex}

\section{Additional Figures}
\label{sec:figures}
\input{./appendix/fig.tex}

\end{document}

%% file: abstract.tex
We theoretically study how differential privacy interacts with both individual and group fairness in binary linear classification. More precisely, we focus on the output perturbation mechanism, a classic approach in privacy-preserving machine learning.  We derive high-probability bounds on the level of individual and group fairness that the perturbed models can achieve compared to the original model. Hence, for individual fairness, we prove that the impact of output perturbation on the level of fairness is bounded but grows with the dimension of the model.  For group fairness, we show that this impact is determined by the distribution of so-called angular margins, that is signed margins of the non-private model re-scaled by the norm of each example.

%% file: intro.tex
With the advent of Machine Learning models as tools that may significantly impact human lives, concerns surrounding their trustworthiness started to arise. Among the several notions of trust considered in the literature, Fairness and Privacy emerged as two very desirable properties. The former seeks for models that do not unjustly discriminate against individuals while the latter aims at protecting the personal information of individuals whose data was used to train the models. While both aspects have been extensively studied in isolation \citep{hardt19, dwork14}, it is only recently that the question of their interactions started to attract some interest \citep{fioretto22survey}. In this paper, we take a step forward in theoretically understanding this interplay.

Fairness arises when the machine learning models directly affect individuals, for example by taking decisions related to their health or by judging whether they should receive a loan. Depending on the problem at hand, the notion of fairness that should be considered changes. Two main families of definitions have emerged in the literature. On the one hand, individual fairness states that similar individuals should be treated similarly \citep{dwork12}. On the other hand, group fairness seeks to prevent discriminatory behaviors against some demographic groups defined by population-level characteristics such as gender or age \citep{calders09,hardt16}. Subsequently, a large amount of the fairness literature has been dedicated to the design of algorithms able to enforce such constraints as can be seen from recent surveys \citep{hardt19,caton2020fairness,mehrabi2021survey,hort2022bia}.
In this work, our objective is not to propose yet another algorithm to learn fair models but, instead, it is to study the theoretical impact of privacy on both individual and group fairness.

The goal of privacy-preserving machine learning is to learn models such that a malicious entity is unable to infer whether the data of an individual took part in model training or not. This objective is most commonly formalized under the notion of differential privacy \citep{dwork14} which bounds the ratio of the probabilities that a mechanism outputs the same sets of models when it learns from two datasets that differ in a single data point. To hide the presence of an individual, differentially private mechanisms use randomization during the training process. For example, in output perturbation, the weights of the learned model are randomized using centered noise \citep{balle18}. Similarly, in noisy gradient descent, the (clipped) gradients are perturbed to obtain a private model \citep{abadi16}. Thus, privacy-preserving mechanisms tend to output models that have equivalent privacy guarantees but are likely to have different outputs on the same set of individuals \citep{kulynych23}. Evaluating the impact of privacy on fairness requires a thorough understanding of the distribution of models that can be returned by a given mechanism and, consequently, an understanding of the distribution of their predictions. In this work, we take a step forward in this direction.

\paragraph{Contributions.} 
We propose an in-depth study of the impact of output perturbation mechanism on fairness in binary linear classification. 
Our contributions are of theoretical nature and span three main concepts that capture different fairness issues, namely individual fairness, disagreement, and group fairness. 
In Section~\ref{sec:individual fairness}, we derive a high-probability bound on the individual fairness of private models compared to the original, non-private model. This bound grows with the dimension of the problem $p$ and the noise parameter of the output perturbation mechanism $\sigma$ at a rate of $O(\sigma\sqrt{p})$. Under additional assumptions, we also show a lower bound on individual fairness that grows at the same rate. It means that the impact of output perturbation on individual fairness, in general, depends on the dimension of the problem.
Second, in Section~\ref{sec:disagreement}, we derive a high probability bound on the prediction disagreement \citep{kulynych23} between a non-private model and its private counterparts. This result shows that the impact of output perturbation on a single individual critically depends on the \emph{angular margin} of the \emph{non-private} model, that is its signed margin divided by the norm of the feature vector. We also extend this result to show that most perturbed models only disagree with the non-private model on a limited number of examples.
Finally, in Section~\ref{sec:bound}, we derive a high-probability bound on the group fairness of the private models compared to the non-private one.  
Our bound grows with the noise parameter $\sigma$  and depends on the distribution of angular margins of the non-private model.

\subsection*{Related Works}

\paragraph{Privacy and predictive multiplicity.} To the best of our knowledge, \citet{kulynych23} were the first to theoretically investigate the problem of predictive multiplicity in privacy. For a fixed example, they study the probability that two models obtained from the same privacy preserving mechanism behave differently. They provide a closed-form expression for this disagreement for the output perturbation mechanism applied to linear models. Interestingly, while they do not mention it explicitly, angular margins, that are at the core of our derivations, appear to play a key role in their results. They also show that the disagreement can be efficiently estimated using a limited number of models drawn from the distribution of private models. Finally, they empirically evaluate the degree of disagreement between models trained using the DP-SGD mechanism \citep{abadi16} and the objective perturbation mechanism \citep{chaudhuri11}. 
In our work, rather than considering the predictive multiplicity between different private models, we focus on the differences in terms of predictions between a non-private model and its private counterparts. Furthermore, beyond the probability of disagreement for a given example, we show that for all but a few private models, the probability that they disagree with the non-private model is bounded by a quantity that depends on the distribution of angular margins of the latter.

\paragraph{Privacy negatively impacts fairness.} \citet{bagdasaryan19} perform  one of the first empirical analyses of the impact of differential privacy on fairness. They show that in image-based gender classification, the expected accuracy of private models trained using DP-SGD \citep{abadi16} is more negatively impacted for darker-skin individuals compared to that of the lighter skin individuals. \citet{farrand20} extend the experimental setting of \citet{bagdasaryan19}, by considering higher class size imbalances on CelebA dataset \citep{celeba15}; they also consider a higher range of privacy levels. In our work, we theoretically analyze output perturbation and show that its impact on group fairness is limited and is controlled by the distribution of angular margins of the non-private model. Note that our results do not contradict these empirical findings since (i) we consider a different privacy preserving mechanism and (ii) we consider linear models instead of complex non-linear neural networks.

\paragraph{Privacy has a limited impact on fairness.} \citet{tran21} and \citet{esipova22} perform one of the first theoretical analyses of the interaction of privacy preserving machine learning with group fairness. The authors study the impact of privacy-preserving mechanism on the expected excess risk gap where the expectation is taken over the randomness of the mechanism. They respectively show that privacy has a limited impact on fairness as long as either the gradients \citep{esipova22} or the Hessians \citep{tran21} are well aligned between different groups. In this paper, we also derive conditions ensuring that privacy has a limited impact on fairness. However, we (i) consider different fairness measures and (ii) obtain different quantities of interest, that is angular margins. The work closest to ours is the one of \citet{mangold22}. It studies the impact of the output perturbation \citep{balle18} and DP-SGD mechanisms \citep{abadi16} on group fairness in classification. They derive high probability bounds showing that the loss of fairness due to privacy decreases at a rate $\tilde O(\sqrt{p\log(1/\delta)}/(\varepsilon n))$, where $p$ is the number of model parameters, $n$ is the dataset size, and $(\varepsilon, \delta)$ are the differential privacy parameters. In this work, we restrict ourselves to linear binary classification problems with output perturbation. It simplifies the analysis and allows us to obtain a bound that is independent of the number of model parameters $p$, improving the sharpness of the result in high-dimensions. Beyond group fairness, we also provide a high probability bound on the loss of individual fairness due to privacy.

%% file: model.tex
We assume that the data can be represented as $\z=(\x,\s,\y)$, where $\x\in\calx\subseteq\R^p$ is a feature space, $\s\in\cals$ is a categorical sensitive attribute such as ethnicity or gender, and $\y\in\caly=\{-1,1\}$ is a binary target label. The training dataset $D=\{\z_i\}_{i=1}^n$ consists of $n$ i.i.d. samples drawn from an unknown distribution $\cald$ over $\calx\times\cals\times\caly$.
We consider the family $\calh$ of linear models $\h_\theta(\x) = \theta^\intercal \x$, where  $\theta\in\R^p$ is the vector of parameters. Given an example $\x \in  \calx$, the prediction $\yh \in \{-1,1\}$ is determined by the sign of the model evaluated at the datapoint, that is $\yh(\h_\theta,\x) = 1$ if  $\h_\theta(\x) \geq 0$ and $\yh(\h_\theta,\x) = -1$ otherwise. Note that we implicitly model the bias term in the model weights by having the last component of each feature vector $\x$ being equal to 1. To simplify the exposition, we sometimes drop the subscript $\theta$ in the model notation $\h_\theta$. Throughout the paper, we use uppercase letters to denote random variables, and lowercase letters to denote their realizations. 

In the remainder of this section, we introduce the notions of accuracy, individual fairness, group fairness, and privacy, that will be used throughout the paper.

\paragraph{Accuracy.}
The accuracy  $\acc$ of a  model $\h$ over $\cald$ is the probability that it makes correct predictions, that is
\begin{align}
    \label{eq:accuracy}
    \acc(\h, \cald) = \Pb_{\Z=(\X,\S,\Y)\sim\cald}\left[\yh(\h,\X)=\Y \right].
\end{align}

\paragraph{Individual fairness.}
We use a standard notion of fairness \citep{mukherjee20} where 
 a model $\h$ is called \emph{$L(\h)$-individually fair} if for any pair of individuals $\x,\x'\in \calx$,
\begin{align}
    \label{eq:individual fairness}
    |\h(\x) - \h(\x')| \le L(\h)\cdot\|\x - \x'\|_2.
\end{align}
We note that any linear model $\h_\theta$ is $\|\theta\|_2$-individually fair, since using Cauchy-Schwarz inequality for any $\x,\x' \in\calx$, we have  $|\theta^\intercal \x - \theta^\intercal \x'|\le\|\theta\|_2 \cdot \|\x-\x'\|_2$. We denote the smallest individual fairness constant $L(\h)$ as  $L^*(\h)$.

\paragraph{Group fairness.} 
We use a unified notation encompassing several group fairness notions \citep{maheshwari2022,mangold22}. It assumes that the set $\calx\times\cals\times\caly$ can be partitioned into $K$ non-intersecting subsets $\tau_k$ (subgroups of the population), that is
$\calx\times\cals\times\caly = \bigcup_{k=1}^K \tau_k.$
The partition is specific to each fairness notion, and typically depends on the value of the sensitive attribute $\S$ and the true label $\Y$. 
The \emph{fairness measure} $\fk$ of a model $\h$ is then defined as
\begin{align}
    \label{eq:fairness}
    \fk(\h,\cald)= C_k^0 + \sum_{k'=1}^K C_{k}^{k'} \cdot \acc(\h,\cald_{k'}),
\end{align}
where $C_k^0, C_k^{k'}$ denote constants that are specific to each fairness notion and independent from model $\h$, and $\cald_k$ denotes the conditional distribution of $\Z=(\X,\S,\Y)$ for the group $k$, that is $\Pb(\Z=\z\mid\Z\in\tau_k)$.

If $\fk(\h,\cald)=0$, then the model is called fair.
If $\fk(\h,\cald)$ is positive then the group $k$ is called \emph{advantaged}, whereas if $\fk(\h,\cald)$ is negative then the group $k$ is called \emph{disadvantaged}. \citet{mangold22} show that classic group fairness notions such as demographic parity \citep{calders09}, equalized odds and equality of opportunity \citep{hardt16}, or accuracy parity \citep{zafar17} can all be represented in the form of \eqref{eq:fairness}.

\paragraph{Privacy.} We consider the standard notion of \emph{differential privacy} \citep[Def.~2.4]{dwork14}.
It states that a randomized mechanism $\calm^\text{priv}$ which takes as input a dataset $D\in(\calx\times\cals\times\caly)^n$ and outputs a prediction model $\h\in\calh$ is $(\varepsilon,\delta)$-differentially private if for any two datasets $D$, $D'$ that differ in a single element, and for any set of models $H\subseteq \mathcal{H}$:
\begin{align*}
    \Pb\left[\calm^\text{priv}(D) \in H\right] \le \exp(\varepsilon) \cdot \Pb\left[\calm^\text{priv}(D')\in H\right] + \delta,
\end{align*}
where $\varepsilon \ge 0$ and $\delta \in [0,1]$.
To guarantee $(\varepsilon,\delta)$-differential privacy, various mechanisms were proposed in the literature. In this paper, we focus on the \emph{output perturbation mechanism} using centered Gaussian noise \citep{balle18}. It consists in perturbing the output of the non-private mechanism $\calm \colon (\calx\times\cals\times\caly)^n\to \calh$ as follows
\begin{align}
    \label{eq:noise model}
    \calm^\text{priv}(D) = \calm(D) + \sigma\cdot\xi,\;\;\;\xi \sim \N(0,\I_p).
\end{align}
In \eqref{eq:noise model}, we implicitly overloaded the ``$+$'' operator: adding a vector $v\in\R^p$ to a linear model $\h_\theta$ parameterized by a weight vector $\theta\in\R^p$ represents a linear model with vector of weights equal to $\theta + v$. The noise parameter $\sigma \ge 0$ controls the level of differential privacy as stated below.
\begin{lemma}[\citet{balle18}]
\label{lemma:dp-guarantee-non-diagonal-covariance} 
The output perturbation mechanism \eqref{eq:noise model} provides $(\varepsilon,\delta)$-differential privacy guarantees if and only if 
    \begin{align*}
        \Phi\left({\Delta}/{2\sigma} - {\varepsilon\sigma}/{\Delta}\right) - \exp(\varepsilon)\cdot\Phi\left(-{\Delta}/{2\sigma}-{\varepsilon\sigma}/{\Delta}\right)\le \delta,
    \end{align*}
    where $\Phi$ denotes the CDF of the standard normal random  variable, $\Delta$ is the sensitivity of the non-private learning mechanism $\calm$ defined as
    $\Delta = \sup_{D,D'} \|\calm(D) - \calm(D')\|_2,$
and $D,D'\in(\calx\times\cals\times\caly)^n$ denote all pairs of datasets that differ in a single element. 
\end{lemma}

%% file: individual_fairness.tex
As our first contribution, we study the impact of output perturbation on  individual fairness. More precisely, given a non-private model $\h$ obtained through the mechanism $\calm$ and its private counterparts $\hp$ obtained using $\calm^\text{priv}$, we prove that the loss of fairness due to privacy is bounded with high probability with respect to the randomness of the privacy-preserving mechanism $\calm^\text{priv}$. The bound grows linearly with the noise parameter $\sigma$, and as $\sqrt{p}$ with the number of model parameters. To achieve this, we start, in the next theorem, by deriving a high-probability bound on the norm of private models $\hp$. This result will serve as the basis to derive our results on individual fairness. We defer all the proofs of our results to Appendix~\ref{sec:proofs}.
\begin{theorem}
\label{theorem:individual fairness}
Let $\calm$ be a non-private mechanism returning $\h$ and $\calm^\text{priv}$ be the output perturbation mechanism \eqref{eq:noise model}.
    \begin{enumerate}
        \item With probability greater than $1-\zeta$ over the randomness of $\calm^\text{priv}$, we have
        \begin{small}
        \begin{align*}
            & \|\theta^\text{priv}\|_2 <  \|\theta\|_2 + \sigma \sqrt{p + 2\sqrt{p\log\left(\frac 1 \zeta\right)} + 2 \log\left(\frac 1 \zeta\right)}.
        \end{align*}
    \end{small}
        \item With probability greater than $1-\zeta$ over the randomness of $\calm^\text{priv}$, we have
        \begin{small}
        \begin{align*}
            &\|\theta^\text{priv}\|_2 > \max \begin{cases}
                \|\theta\|_2 - \sigma \sqrt{p + 2\sqrt{p\log\left(\frac 2 \zeta\right)} + 2 \log\left(\frac 2 \zeta\right)},\\
                {\sigma}\sqrt{\max\left(0, p - 2\sqrt{p\log\left(\frac 2 \zeta\right)}\right)} - \|\theta\|_2.
            \end{cases}
        \end{align*}
        \end{small}
\end{enumerate}
\end{theorem}

In the general setting, $L^*(\hp)\le \|\theta^\text{priv}\|_2$, hence, only the upper bound from Theorem~\ref{theorem:individual fairness} can be applied to bound the individual fairness constant of private models $L^*(\hp)$. If $\calx$ is an open set, then we show in Appendix~\ref{proof:individual fairness} that $L^*(\hp) =\|\theta^\text{priv}\|_2$, so both lower bound and upper bound from Theorem~\ref{theorem:individual fairness} can be used.

To summarize, we prove that the \emph{upper bound on the individual fairness} of private models  $L(\hp)$ grows with the dimension $p$ and the noise parameter $\sigma$ as $O(\sigma \sqrt{p})$. For empirical risk minimization problems with convex loss functions and strongly-convex penalty terms \citep{chaudhuri11}, for a fixed privacy parameter $\varepsilon$, the noise parameter  $\sigma$ is decreasing with the number of training examples $n$ at a rate of $1/n$.  Hence, the above result  suggests that collecting more data can reduce the impact of differential privacy on the individual fairness constant $L(\hp)$. 

Under the additional assumption that $\calx$ is an open set, we also show a lower bound on the individual fairness constant which grows with the noise $\sigma$ and the dimension $p$ hinting that the dependence on these two quantities is unavoidable. As we will show in Section~\ref{sec:bound}, this contrasts with our results for \emph{group fairness} notions for which we show that the impact of output perturbation is dimension-independent.

\begin{figure}
    \centering
    \begin{tikzpicture}[scale=0.22/.3]
        \node (img)  {\includegraphics[width=.4\linewidth]{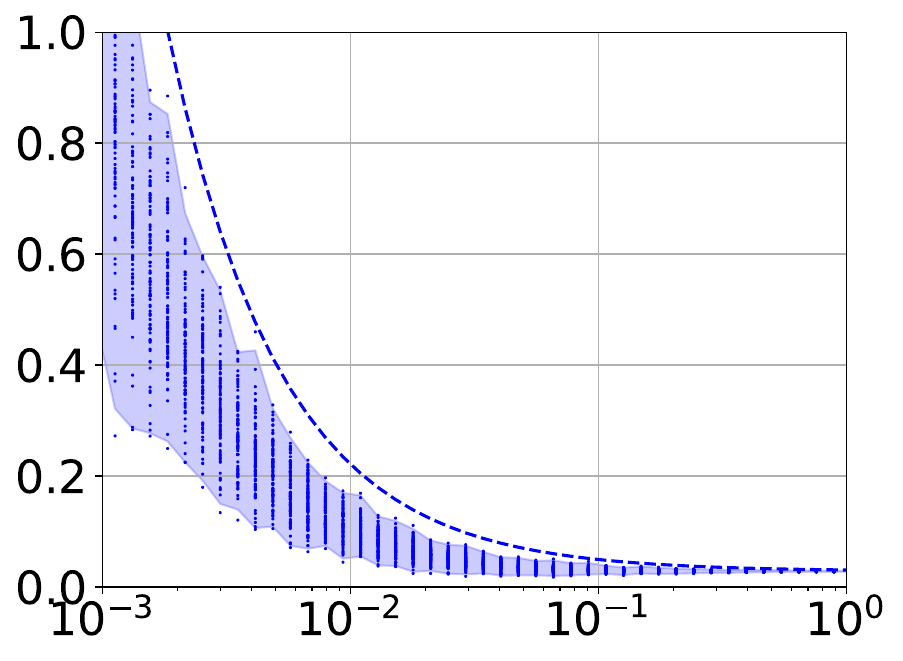}};
        \node[below=of img, node distance=0cm, yshift=4em,font=\color{black}] {$\varepsilon$};
        \node[left=of img, node distance=0cm, rotate=90, anchor=center,yshift=-2.5em,font=\color{black}] {$\|\theta^\text{priv}\|_2$};
    \end{tikzpicture}
    \caption{Individual fairness  of perturbed models $\hp$ on \texttt{Adult} dataset. The $99\%$-confidence bounds are shown by dashed lines, and color-filled regions correspond to regions where $99\%$ of measurements lie.}
    \label{fig:individual fairness}
\end{figure}

We illustrate the individual fairness upper bound from Theorem~\ref{theorem:individual fairness} in Fig.~\ref{fig:individual fairness} for the \texttt{Adult} dataset \citep{dua17}. The non-private model $\h$ is obtained by training the $l_2$-regularized logistic regression with penalty term $C=1$ on $80\%$ of the data. For each value of the privacy parameter $\varepsilon$, we generate $100$ private models $\hp$. To calculate the confidence bounds, we use the remaining $20\%$ of the data. The random seed is set to $0$ for all figures (see Appendix~\ref{sec:figures} for other seeds). We plot the norm of each private model weights $\|\theta^{\text{priv}}\|_2$ as a single point. The high probability bound from Theorem~\ref{theorem:individual fairness} is shown as a dashed line. We observe that the bound is tight for the chosen range of privacy parameters $\varepsilon$ and fixed $\delta=1/n^2$.

In this section, we have shown that a private model derived from a model that tends to treat close individuals similarly will also treat close individuals similarly. However, this does not tell us how the individuals themselves will be impacted by privacy. This is what we study in the next section.

%% file: disagreement.tex
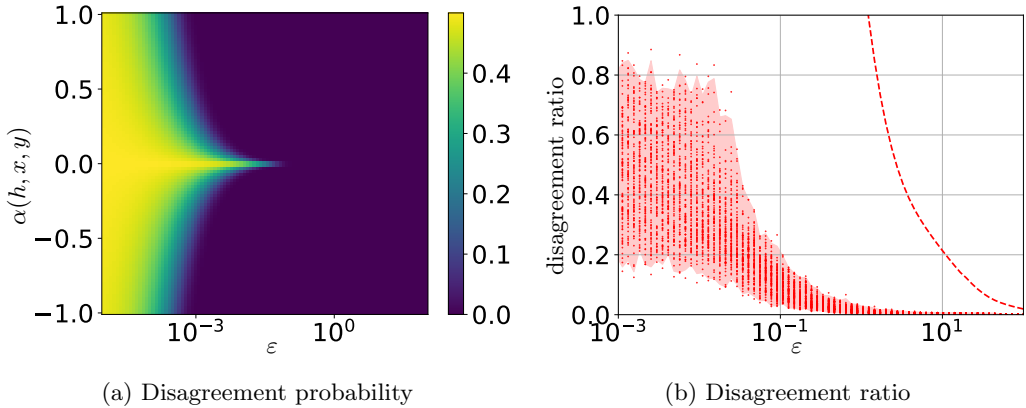
\begin{figure}
    \centering
    \begin{subfigure}{0.45\linewidth}
        \centering\begin{tikzpicture}[scale=0.22/.3]
            \node (img)  {\includegraphics[width=.95\linewidth]{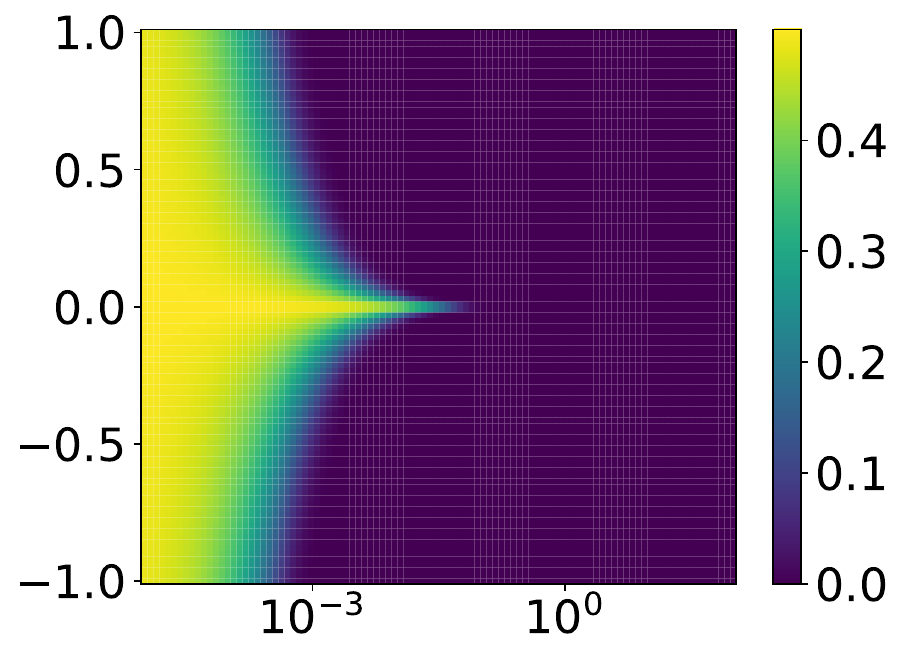}};
            \node[below=of img, node distance=0cm, yshift=4em,font=\color{black}] {\small$\varepsilon$};
            \node[left=of img, node distance=0cm, rotate=90, anchor=center,yshift=-3em,font=\color{black}] {\small$\amargin(\h,\x,\y)$};
        \end{tikzpicture}
        % \caption{}
        \caption{Disagreement probability}
        \label{fig:disagreement probability}
    \end{subfigure}
    \begin{subfigure}{.45\linewidth}
        \centering
        \begin{tikzpicture}[scale=0.22/.3]
            \node (img)  {\includegraphics[width=.9\linewidth]{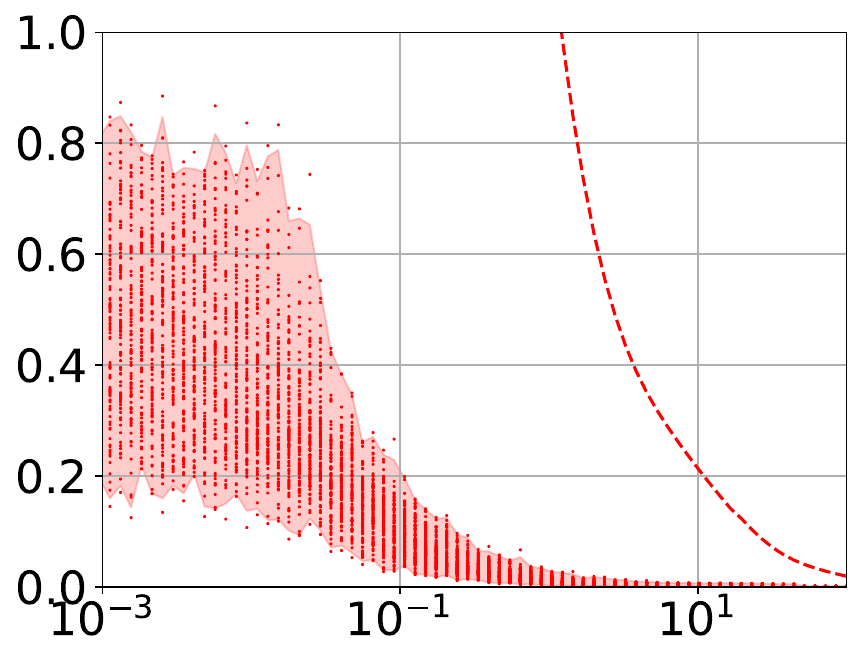}};
            \node[below=of img, node distance=0cm, yshift=4em,font=\color{black}] {\small $\varepsilon$};
            \node[left=of img, node distance=0cm, rotate=90, anchor=center,yshift=-3em,font=\color{black}] {\small disagreement ratio};
        \end{tikzpicture}
        % \caption{}
        \caption{Disagreement ratio}
        \label{fig:disagreement ratio}
    \end{subfigure}
    \caption{(a) Disagreement probability of the perturbed model $\hp$ with the unperturbed model $\h$ for a data point; (b) Disagreement ratio of perturbed models $\hp$ with the unperturbed model $\h$  on the \texttt{Adult} dataset. On panel (b), the $99\%$-confidence bounds are shown by dashed lines, and color-filled regions correspond to regions where $99\%$ of measurements lie.}
    \label{fig:disagreement}
\end{figure}

As our second contribution, we study the level of disagreement between a non-private model $\h$ obtained through the mechanism $\calm$, and its private counterparts $\hp$ obtained using output perturbation $\calm^\text{priv}$. It corresponds to cases where their predictions differ due to the randomization in the mechanism $\calm^\text{priv}$. 
Hence, in our first lemma, we compute the probability that the private models disagree with the non-private one for a single example $\x \in \calx$. This probability is a decreasing function of the absolute value of the so-called \emph{angular margin $
\amargin(\h,\x,\y)$}, 
that is confidence margins scaled by the norm of the feature vector as 
$$\amargin(\h,\x,\y) = {\rho(\h,\x,\y)}/{\|x\|_2},$$
where the quantity $\rho(\h,\x, \y)= \y \cdot \h(\x)$ is called the \emph{signed margin} of the model $\h$ for the data point $(\x,\y)$ \citep{allwein01}. Its absolute value, that is $|\rho(\h,\x,\y)|$, is a quantification of the confidence of a model in its predictions, with larger values indicating higher confidence.
\begin{lemma}
    \label{lemma:disagreement per data}
    The probability that the private models $\hp$ drawn from $\calm^\text{priv}$ disagree with the non-private model $\h$ for a given $\x\in\calx$ is
    \begin{align*}
        \Pb_{\hp}\left[\yh(\hp,\x) \not = \yh(\h,\x)\right] = \Phi\left(\frac{-\left|\amargin(\h,\x,\y)\right|}{\sigma}\right),
    \end{align*}
    where $\Phi$ is the CDF of the standard normal distribution.
\end{lemma}
Note that the above result measures the disagreement probability of the private model $\hp$ with the non-private model $\h$, whereas \citet{kulynych23} quantify the disagreement probability between two private models $\hp_1$ and $\hp_2$  that can be obtained using output perturbation from $\h$.
In Fig.~\ref{fig:disagreement probability}, we illustrate the disagreement probability for different values of angular margin $\amargin$ and different values of the privacy parameter $\varepsilon$ and $\delta\approx10^{-9}$. We can identify three main regimes. For large $\varepsilon$, the probability that $\hp$ disagrees with the non-private model $\h$ is small. For intermediate $\varepsilon$, the data points with lower angular margins have higher disagreement probability. For small $\varepsilon$, the disagreement probability reaches the value of $0.5$ for all the examples.

We now aim to quantify the proportion of such disagreements across the overall data distribution $\cald$, that is to bound the proportion of data points $\x$ on which a private model $\hp$ disagrees with the non-private model $\h$. Theorem~\ref{theorem:disagreement number} tells us that, with high probability over the randomness of the privacy-preserving mechanism $\calm^\text{priv}$, this proportion is bounded by the expectation over data distribution $\cald$ of the disagreement probability. Therefore, the disagreement ratio tends to be smaller if the angular margins of the \emph{non-private model $\h$} are large in absolute values.
\begin{theorem}[Disagreement ratio bound]
    \label{theorem:disagreement number}
    With probability greater than $1-\zeta$ over the randomness of  $\calm^\text{priv}$, the disagreement ratio of $\hp$ is bounded
    \begin{align*}
        \Pb_\cald\left[\yh(\hp,\X) \not = \yh(\h,\X)\right] < \frac{\E_\cald\left[\Phi\left(\frac{-\left|\amargin(\h,\X,\Y)\right|}{\sigma}\right)\right]}{\zeta}.
    \end{align*}
\end{theorem}

In Fig.~\ref{fig:disagreement ratio} we plot the disagreement ratio for different privacy parameters $\varepsilon$ for the \texttt{Adult} dataset \citep{dua17}.   We use the same procedure as in Fig.~\ref{fig:individual fairness} for the private and non-private models. Our  99\%-confidence bound on the disagreement ratio is non-trivial for values of $\varepsilon \in (1, 10)$. 

While measuring disagreement provides some insights on how close the predictions of the private and non-private models are, it is difficult to directly connect it to group fairness. Indeed, it does not tell us anything about the behavior of the models among groups, that is a higher disagreement for one sensitive group does not necessarily mean that group fairness will be negatively impacted. It means that a more specific analysis is needed. This is the goal of the next section.

%% file: group_fairness.tex
In this section, we study the impact of output perturbation on group fairness. In Section~\ref{ssec:expected fairness}, we bound the expectation and the variance of the fairness measure $\calf_k(\hp)$ with respect to the randomness in the privacy preserving mechanism $\calm^\text{priv}$. Interestingly, both quantities strongly depend on the distribution of angular margins $\amargin(\h,\x,\y)$ of the non-private model, a concept that was already key in the measurement of disagreement. Then, using the expressions on expectation and variance of fairness measures, in Section~\ref{ssec:bound} we derive a high probability bound on the group fairness of a given private model.

\subsection{Expectation and Variance of Group Fairness}
\label{ssec:expected fairness}

We start by showing that the expectation of both the accuracy $\cale_\acc=\E_{\hp} \left[\acc (\hp, \cald)\right]$ and the group fairness $\cale_{\calf_k} =\E_{\hp} \left[\fk (\hp, \cald)\right]$  over the randomness of the privacy-preserving mechanism $\calm^\text{priv}$ can be written in terms of the expectation over the distribution of angular margins $\amargin$ of the non-private model $\h$. 
\begin{lemma}[Expected fairness of output perturbation]
    \label{lemma:first moment}
    Let $\calm$ be a non-private mechanism returning $\h$ and $\calm^\text{priv}$ be the output perturbation mechanism \eqref{eq:noise model}.
    \begin{enumerate}
        \item The expected accuracy $\acc$ over the randomness of private models $\hp$ equals
        \begin{align}
            \label{eq:expected accuracy}
            \cale_\acc(\h,\sigma,\cald) = \E_{\cald}\left[\Phi\left({\amargin(\h,\X,\Y)}/{\sigma}\right)\right].
        \end{align}
        \item The expected fairness $\fk$ over the randomness of private models $\hp$ equals
        \begin{align}
            \label{eq:expected fairness}
            \cale_{\calf_k}(\h,\sigma,\cald) = C_k^0 + \sum_{k'=1}^K C_k^{k'}\cale_{\acc}(\h,\sigma,\cald_{k'}).
        \end{align}
    \end{enumerate}
\end{lemma}

From the previous result alone, one could be tempted to conclude that as the right hand side of Equation \eqref{eq:expected fairness} becomes $0$, for example when the noise becomes large while $C_k^0=0$ and $\sum_{k'=1}^K C_{k}^{k'}=0$ as in accuracy parity \citep{mangold22}, the private models tend to be fair. However, this conclusion would be erroneous. Indeed, $\calf_k \in [-1,1]$ and thus it could be that the fairness of private models that advantage and disadvantage a given group compensate one another. Thus, the previous result alone is not sufficient to conclude anything on the group fairness of private models and we also need to look at the variance to get a more complete picture. This is done in the next lemma where we upper bound it, showing that angular margins are, once again, the key quantities that need to be considered.
\begin{lemma}[Variance of fairness of private models $\hp$]
    \label{lemma:variance of fairness bound}
    Let $\calm$ be a non-private mechanism returning $\h$ and $\calm^\text{priv}$ be the output perturbation mechanism \eqref{eq:noise model}.
    \begin{enumerate}
        \item The variance of the accuracy $\acc$ over the randomness of the privacy-preserving mechanism $\calm^\text{priv}$ is bounded from above, that is
        \begin{small}
            \begin{align*}
             &\V_{\hp}\left[\acc(\hp,\cald)\right]\le \calv_{\acc}(\h,\sigma, \cald) = {\mathop{\E}_{\Z^{(1)},\Z^{(2)}\sim \cald}\left[\Phi\left(\frac{\amargin(\h,\Z^{\text{min}}}{\sigma}\right) \Phi\left(\frac{-\amargin(\h,\Z^{\text{max}})}{\sigma}\right)\right]},
            \end{align*}
        \end{small}
        where 
        $\Z^{\text{min}} = \argmin_{\Z^{(1)}, \Z^{(2)}} \amargin(\h,\Z)$ and  $\Z^{\text{max}} = \argmax_{\Z^{(1)}, \Z^{(2)}} \amargin(\h,\Z)$.
    \item The variance of the group fairness measure $\calf_k$ over the randomness of the privacy-preserving mechanism $\calm^\text{priv}$ is bounded from above, that is
    \begin{small}
        \begin{align*}
            &\V_{\hp}\left[\fk(\hp,\cald)\right] \le \calv_{\fk}(\h,\sigma, \cald),
        \end{align*}
    \end{small}
    where $\calv_{\fk}=\left[\sum_{k'=1}^K |C_{k}^{k'}| \sqrt{\calv_{\acc}(\h,\sigma, \cald_{k'})}\right]^2$.
    \end{enumerate}
\end{lemma}

It is worth noting that the smaller is the noise parameter $\sigma$, the lower is the fairness variance. Indeed, we either have $\lim_{\sigma\to 0}\Phi(\amargin(\h,\Z^{\text{min}})/\sigma)=0$ or $\lim_{\sigma\to 0} \Phi(-\amargin(\h,\Z^{\text{max}})/\sigma)=0$. This is an expected but desirable behaviour for a variance upper bound since, in this case, the private models tend to be identical to the non-private model and thus tend to have similar fairness levels. Similarly, as the noise becomes large, the right hand side becomes $\frac{1}{2}\sum_{k'=1}^K |C_{k}^{k'}|$ which might be close to $1$. For example, for accuracy parity it holds that $\sum_{k'=1}^K |C_{k}^{k'}| \leq 2$ \citep{mangold22}. In this case, anything may happen as imposing privacy could lead to both models that strongly advantage or disadvantage specific groups.

\subsection{High Probability Bound on Group Fairness}
\label{ssec:bound}

In the previous section, we derived bounds on the expectation and variance of group fairness given the randomness in the privacy preserving mechanism. In this section, we show that, using Chebyshev's inequality, we can derive bounds on the accuracy and fairness of individual models $\hp$ that hold with high probability. This is summarized in the next theorem for fairness and in Appendix~\ref{proof:confidence bounds} for accuracy.
\begin{theorem}\label{theorem:confidence bounds}
    Let $\cale_{\fk}$ and $\calv_{\fk}$ denote the expected fairness and the fairness variance  upper bound defined in Lemma~\ref{lemma:first moment} and  Lemma~\ref{lemma:variance of fairness bound}. With probability at least $1-\zeta$ over the randomness of $\calm^\text{priv}$, we have
    \begin{align*}
        \begin{cases}
            \fk(\hp,\cald)  <  \cale_{\fk}(\h,\sigma,\cald) + \sqrt{\frac{\calv_{\fk}(\h,\sigma,\cald)}{\zeta}}, \\
            \fk(\hp,\cald)  > \cale_{\fk}(\h,\sigma,\cald) - \sqrt{\frac{\calv_{\fk}(\h,\sigma,\cald)}{\zeta}}.
        \end{cases}
    \end{align*}
\end{theorem}

The main takeaway of this theorem is that non-private models with angular margins that are far from $0$ tend to be less impacted by output perturbation. Furthermore, we emphasize that the right hand side of the bound depends only on the distribution of angular margins of the non-private model $\h$, the data distribution $\cald$, the noise parameter $\sigma$, and the confidence parameter $\zeta$. In particular, it does not depend on $p$ the number of parameters of the model. To more precisely position this result with respect to the state of the art, we propose to compare it to the bound of \citet{mangold22} that was derived for models with Lipschitz-continuous margins and is recalled below.
\begin{theorem}[\citet{mangold22}]
\label{theorem:mangold}
    With probability greater than $1-\zeta$ over the randomness of $\calm^\text{priv}$, we have
    $$\begin{cases}
            \fk(\hp,\cald) \le \fk(\h,\cald) +  P(\h,\sigma,\cald,\zeta), \\
            \fk(\hp,\cald) \ge \fk(\h,\cald) -P(\h,\sigma,\cald,\zeta),
    \end{cases}$$
    where {\small $P =
        \sum_{k'=1}^K |C_k^{k'}| \Pb_{\cald_{k'}}\left[\frac{|\rho(\h,\X,\Y)|}{L_{\X,\Y}}  \le 2\sigma\sqrt{\log\left(\frac 2 \zeta\right)p}\right]$}, $L_{\X,\Y}$ is the Lipschitz constant of the margin $\rho(\h,\X,\Y)$ with respect to $\h$, and $p$ is the number of model parameters.
\end{theorem}
The main disadvantage of our bound lies in the dependence in $\zeta$ where our bound is slightly worse with a rate of $1/\sqrt{\zeta}$ compared to $\sqrt{\log(1/\zeta)}$. However, the bound in Theorem~\ref{theorem:mangold} grows unbounded with the number of model parameters as $\sqrt{p}$ appears in the right hand side while, as mentioned before our bound does not depend on the dimension of the model. Hence, our result seems more applicable for larger models.  Interestingly, we also note that despite the different proof techniques used to obtain the bounds, there is a close relation between the quantity $\rho(\h,\X,\Y) / L_{\X,\Y}$ in Theorem~\ref{theorem:mangold} and the notion of angular margin $\amargin(\h,\X,\Y)$. Indeed, for linear models, the two values coincide.

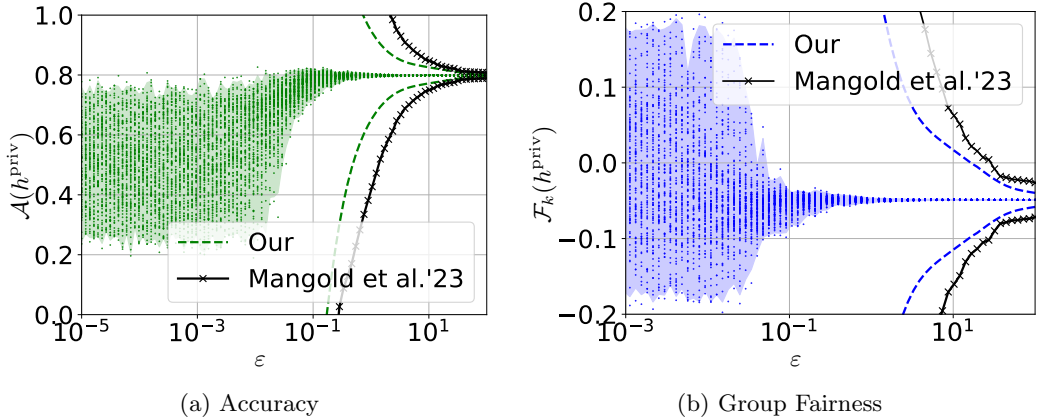
\begin{figure}
    \centering
    \begin{subfigure}{.45\textwidth}
        \centering
        \begin{tikzpicture}[scale=0.22/.3]
            \node (img)  {\includegraphics[width=0.9\linewidth]{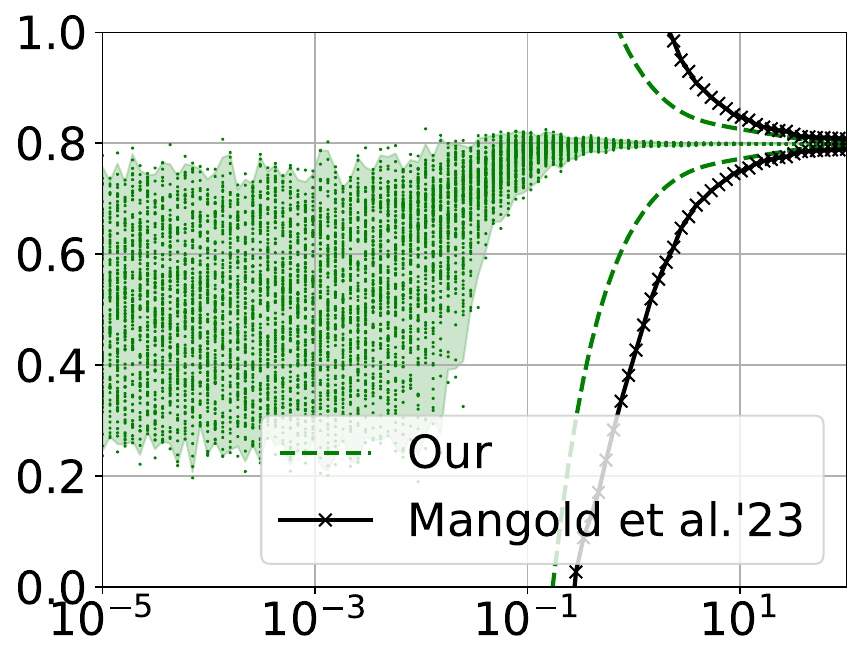}};
            \node[below=of img, node distance=0cm, yshift=3.5em,font=\color{black}] {\small $\varepsilon$};
            \node[left=of img, node distance=0cm, rotate=90, anchor=center,yshift=-3em,font=\color{black}] {\small $\acc(\hp)$};
        \end{tikzpicture}
        \caption{Accuracy}
        \label{fig:accuracy}
    \end{subfigure}
    \begin{subfigure}{.45\textwidth}
        \centering
        \begin{tikzpicture}[scale=0.22/.3]
            \node (img)  {\includegraphics[width=0.95\linewidth]{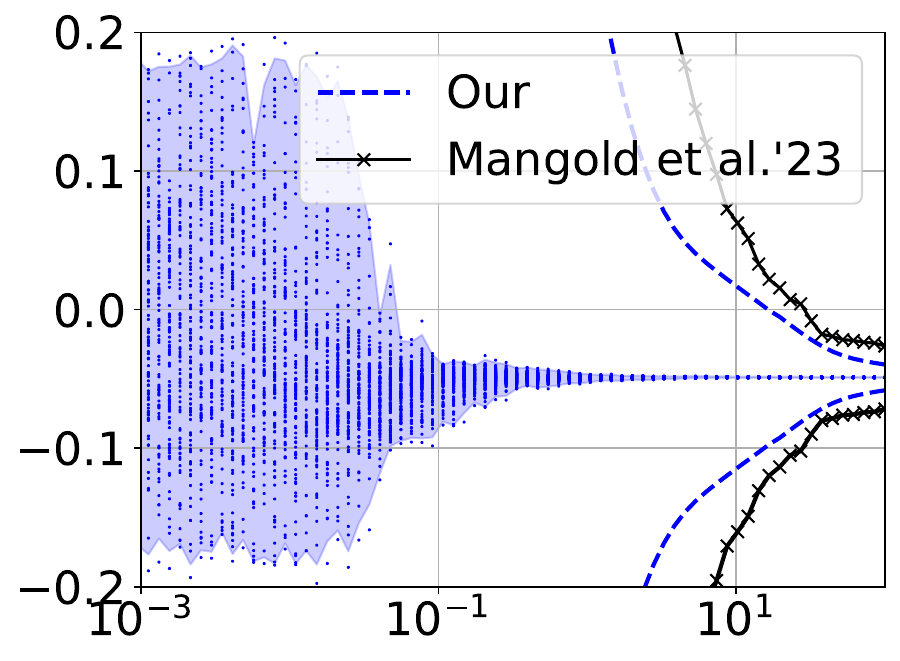}};
            \node[below=of img, node distance=0cm, yshift=3.5em,font=\color{black}] {\small$\varepsilon$};
            \node[left=of img, node distance=0cm, rotate=90, anchor=center,yshift=-3em,font=\color{black}] {\small$\calf_k(\hp)$};
        \end{tikzpicture}
        \caption{Group Fairness}
        \label{fig:fairness}
    \end{subfigure}
    \caption{Accuracy $\acc$ and  group fairness $\fk$ (accuracy parity) of private models  $\hp$ for different values of $\varepsilon$ on \texttt{Adult}. 
    The $99\%$-confidence bounds are shown by dashed and crossed lines. 
    The color-filled regions are the ones where $99\%$ of measurements lie.
    }
    \label{fig:bounds}
\end{figure}

In Fig.~\ref{fig:bounds}, we illustrate our bounds on the \texttt{Adult} dataset using the same setting as in Fig.~\ref{fig:individual fairness}. We treat the gender of an individual as the sensitive attribute. We plot the accuracy and fairness of each private model $\hp$ by a single point. We depict our bounds on accuracy (Fig.~\ref{fig:accuracy}) and fairness (Fig.~\ref{fig:fairness}) using dashed lines and the corresponding bounds from \citet{mangold22} with crossed lines. We  observe that, in this setting, our bound is tighter even though we set the confidence parameter to be as small as $\zeta=0.01$.

So far, we have shown that the impact of output perturbation on various fairness quantities is bounded. In the next section, we show that our theoretical results could be relevant in a variety of different settings.

%% file: various-settings.tex
In this section, we analyze how the bounds on individual fairness, disagreement, and group fairness from Sections~\ref{sec:individual fairness},~\ref{sec:disagreement}, and~\ref{sec:bound} can be used in various scenarios. In Section~\ref{sec:bayesian auditing}, we show how the bounds can be used to verify whether a model $\h$ satisfies a promised level of fairness without accessing it directly but via observing a private version $\hp$ of $\h$ and the noise parameter $\sigma$. 
Next, in Section~\ref{ssec:noisy sgd}, we discuss that, under some assumptions used in the literature to study the dynamics of gradient descents \citep{mandt17}, our results can also be applied to Noisy GD, a relaxation of DP-SGD \citep{abadi16} which is another popular privacy-preserving mechanism.
Finally, in Section~\ref{sec:finite sample}, we derive a finite-sample bound for our results on group fairness.

\subsection{Auditing Using Private Models}
\label{sec:bayesian auditing}

We first show how our bounds can be used to assess a model's fairness without accessing it directly. We assume that the auditor has some prior knowledge about the unknown model $\h$. We model this prior as a multivariate normal distribution. The auditor observes a noisy model $\hp$ obtained from the model $\h$ using the output perturbation mechanism \eqref{eq:noise model}. In the next lemma, we show that the posterior distribution of weights of the model $\h$ given its private release $\hp$ is a multivariate normal distribution. It is a corollary of the classic result on the conditional distribution of components of multivariate Gaussian vector \citep[Theorem~4.4d]{rencher08}.
\begin{lemma}
    \label{theorem:bayesian estimate}
    Let $\hp$ be the private model  obtained from the model $\h$ using the output perturbation mechanism \eqref{eq:noise model}. Assume a Bayesian prior on model weights  $\Theta \sim \N(\mu, \eta^2\I_p)$. Then, the posterior distribution of model weights $\h$ follows a multivariate Gaussian distribution 
    \begin{small}
    \begin{align*}
        &\Theta \mid \Theta^\text{priv}=\theta^\text{priv}, \sigma \sim \N\left(\mu +\left[1+\frac{\sigma^2}{\eta^2}\right]^{-1}(\theta^\text{priv}-\mu),\left[1 + \frac{\sigma^2}{\eta^2} \right]^{-1}\sigma^2\I_p\right).
    \end{align*}
    \end{small}
    In particular, for the uniform prior, that is $\eta^2 \to \infty$, we have $\Theta \mid \Theta^\text{priv}=\theta^\text{priv}, \sigma \sim \N\left(\theta^\text{priv},\sigma^2\I_p\right)$.
\end{lemma}

This lemma shows that to evaluate the original model $\h$ for fairness, we can again consider the output perturbation mechanism, where the weights of the private model $\theta^\text{priv}$ are perturbed by centered noise that only depends on $\sigma$ and $\eta$. Furthermore, with a uniform prior, it boils down to considering
$\theta = \theta^\text{priv} + \sigma \cdot \xi$, where $\xi\sim\N(0,\I_p)$.
Hence, all the bounds we derived in the previous sections can be applied by simply exchanging the role of $\h$ and $\hp$ in Theorems~\ref{theorem:individual fairness},~\ref{theorem:disagreement number}, and~\ref{theorem:confidence bounds}.
It implies that with high-probability, we can bound individual fairness, group fairness, and accuracy of the non-private model. Furthermore, if the auditor has some prior knowledge about the non-private model, it can leverage it to obtain tighter upper bounds. Note that our assumption requires some knowledge about the noise generation process used in the privacy mechanism employed to obtain $\hp$ which can potentially lead to privacy leaks.

\subsection{The Noisy GD Mechanism}
\label{ssec:noisy sgd}

All our results so far assume that output perturbation was used to enforce privacy.
In this section, we show that, under some assumptions, our bounds can also be used for another privacy mechanism called \emph{noisy gradient descent}.

The noisy GD mechanism is a relaxation of the DP-SGD mechanism \citep{abadi16} which has been designed to enforce privacy in empirical loss minimization based on gradient descent. The main idea behind DP-SGD is that, at each optimization step, the gradient $\nabla_\theta \widehat\call$ of the loss $\widehat\call$ is clipped and perturbed. This allows for a finer control of the overall added noise. In this section, we assume that the sensitivity of the loss gradient is fixed and known to the
practitioner and thus we can ignore the clipping step. We also consider that the full gradient is used in each step. To obtain a private model $\hp$, at each iteration $t=1,\dots, T-1$, the model weights are updated using the perturbed gradient:
\begin{align}
    \label{eq:noisy sgd}
    \theta^{(t+1)} = \theta^{(t)} - \eta \left[\nabla_\theta\mathcal{\widehat L}(\theta^{(t)}, D) + \sigma \cdot \xi^{(t)}\right],
\end{align}
where $D$ is the training data, $\eta$ is the learning rate, and the random noise follows a normal distribution $\xi^{(t)}\sim\N(0,\I_p)$.

In the rest of this section, we show that the stationary distribution of model weights obtained using noisy GD can be approximated by a multivariate normal distribution with mean $\theta^*$  (the optimal non-private model), and a covariance matrix which depends on noise parameter $\sigma$, the learning rate $\eta$, and the Hessian of the loss function at the optimum $H$. This implies that noisy GD can be seen as perturbation of $\theta^*$ with non-isotropic  Gaussian noise.  Interestingly, our results from the previous sections also hold in this case as can be seen in the Appendix where we prove more general results than the ones displayed in the main paper. Thus, the fairness levels of private models obtained through the noisy GD mechanism can also be bounded.

\paragraph{Approximating the dynamics of Noisy GD.} 
The assumptions, inspired by that of  \citet{mandt17} and \citet{koskela23}, are stated below.
\paragraph{(A1)} The empirical loss is well-approximated by a quadratic function, that is $\mathcal{\widehat L}(\theta) = \mathcal{\widehat  L}(\theta^*) + \frac 1 2 (\theta-\theta^*)^\intercal H (\theta-\theta^*),$
where $H_{ij}=\frac{\partial^2_\theta \mathcal{\widehat L}(\theta)}{\partial\theta_i\partial\theta_j}|_{\theta=\theta^*}$ is the symmetric positive definite Hessian at the optimum $\theta^* = \arg\min_{\theta\in\R^p}\mathcal{\widehat L}(\theta)$. 
\paragraph{(A2)} The noisy GD dynamics is well-approximated by its continuous-time dynamics
\begin{align*}
    d\theta_t = -\eta \left(H (\theta_t-\theta^*)dt +  \sigma \I_p dW_t\right),
\end{align*}
where $dW_t$ is the Wiener process. 

An implication of the  above assumptions is the following result on the stationary distribution of model weights of noisy GD mechanism. This is a direct corollary of the result on the stationary distribution of Ornstein-Uhlenbeck process, for example, see \citep{godreche19}.
\begin{lemma}
    \label{lemma:noisy sgd}
    Under assumptions (A1)--(A2), the stationary distribution of the model weights of noisy GD mechanism is multivariate Gaussian, that is
        $\theta^\text{priv} \sim \N\left(\theta^*,\frac 1 2 {\sigma^2\eta}   H^{-1}\right).$
\end{lemma}
In other words, noisy GD can approximately be seen as a case of the perturbation with Gaussian noise with covariance matrix $\Sigma=\frac 1 2 \eta H^{-1}$ and the noise parameter $\sigma$.

\paragraph{Discussion on the assumptions.}
The assumptions (A1)-(A2) are relatively strong and whether they hold or not has been tested in the prior literature. Hence, \citet{hyland19} provide an empirical estimation of the distribution of stochastic gradient descent (without perturbation and clipping), and show that it can be well-approximated by a Gaussian distribution for the case of convex loss functions. However, they also show that for non-convex problems such as the ones involving neural networks, the asymptotic normality does not hold. Similarly, \citet{maddox19} show that assumption (A1) does not hold for deep neural networks in general. In their numerical analysis, they show that in this case the Hessian at the convergence point is not positive definite. From a theoretical standpoint, \citet{zaiwei22} provide the conditions on the asymptotic normality of noisy GD with constant stepsize for smooth and strongly-convex objectives.

\subsection{Finite Sample Analysis of Group Fairness}
\label{sec:finite sample} 

In Theorem~\ref{theorem:confidence bounds} we assume that the  whole data distribution $\cald$ is available. Interestingly, the result also holds for an empirical distribution induced by a finite dataset $D$. That is, with probability greater than $1-\zeta$ over the  randomness of private models $\hp$, we have $$\left|\fk(\hp,D) - \cale_{\fk}(\h,\sigma,D) \right| < \sqrt{\calv_{\fk}(\h,\sigma,D) / \zeta}.$$
In practice, however, we only observe a sample $D$ of size $n$ which is drawn from another unknown distribution $\cald$ and we need to bound the fairness evaluated on the finite sample to the one on the overall distribution. This is done in the next Lemma where we bound the  group fairness of private models on the true distribution $\cald$ using the empirical fairness of the non-private model. It is a corollary of Lemma~3.4 from \citet{mangold22}.
\begin{lemma}
    \label{theorem:finite sample}
    Assume that $n \ge \frac{8\log((2K+1) / \kappa)}{\min_{k'} p_{k'}}$ where $p_{k'}$ is the true proportion of examples from group $k'$. Assume also that $\Pb_{D}\left(\sum_{k'=0}^K\left|C_k^{k'} - \widehat C_{k}^{k'}\right|>\alpha_C\right)\le B_3 \exp(-B_4 \alpha_C^2n)$.
    With probability $1-\kappa$ over the randomness of dataset $D$ of size $n$, with probability $1-\zeta$ over the randomness of private models $\hp$,  we have  
    \begin{small}
    \begin{align*}
        &\left|\fk(\hp,\cald) - \cale_{\fk}(\h,\sigma,D)\right| \le \sqrt{{\calv_{\fk}(\h,\sigma,D)}/{\zeta}} + O\left(\sum_{k'=1}^K |\widehat C_k^{k'}|\sqrt{\frac{d_{\calh} + \log(K/\kappa)}{n p_{k'}}}\right),
    \end{align*}
    \end{small}
    where $d_{\calh}$ is the Natarajan dimension of the class of linear models $\calh$.
\end{lemma}
This theorem thus shows that our bounds are applicable in the finite sample setting up to an error which decreases with the size $n$ of the dataset  but grows with the number of model parameters $p$ through its Natarajan dimension $d_{\calh}$.

%% file: conclusion.tex
In this paper, we theoretically investigated the impact of output perturbation on individual fairness,  disagreement, and group fairness in binary linear classification.  We showed that such impact can be bounded and that a key quantity of interest is the distribution of angular margins of the non-private model.

\paragraph{Limitations and perspectives.} An obvious limitation of our results is that we only consider binary linear classification. It would be interesting to consider non-linear methods, for example by leveraging the rich literature on translation-invariant kernels approaches that can be seen as learning linear models in a space induced by random projections \citep{chaudhuri11}. Similarly, while we bound the loss of fairness due to privacy, we do not address the problem of learning fair and private models. An interesting future work could thus be to leverage our findings on angular margins to create new algorithms.

%% file: appendix/proofs.tex
In this section, we provide proofs of all results stated in the main text. We consider a more general case of perturbation than that in \eqref{eq:noise model}. It  consists in perturbing the output of the non-private mechanism $\calm \colon (\calx\times\cals\times\caly)^n\to \calh$ using Gaussian noise with a possibly non-diagonal positive-definite covariance matrix $\Sigma$, that is
\begin{align}
    \label{eq:general noise model}
    \calm^\text{priv}(D) = \calm(D) + \sigma\cdot\xi,\;\;\;\xi \sim \N(0,\Sigma),
\end{align}
We note that the matrix $\Sigma$ might also depend on the dataset $D$, hence, the mechanism \eqref{eq:general noise model} does not have the same privacy guarantees as the standard output perturbation mechanism \eqref{eq:noise model}. The privacy analysis of such mechanism with data-dependent covariance matrix $\Sigma(D)$ is an interesting research question, however, it is beyond the scope of this paper.

\subsection{Technical Lemmas}

In this section, we state several results from linear algebra and probability theory that are used in the proofs.

We use the following lemma to bound the quadratic forms.
\begin{lemma}
\label{lemma:quadratic form}
Let $A \in \R^{p\times p}$ be a symmetric matrix. Then, for all vectors $\x\in\R^p$, we have:
\begin{align*}
\lambda^{\min}\|\x\|_2^2 \le \x^\intercal A\x \le  \lambda^{\max}\|\x\|_2^2,
\end{align*}
where $\lambda^{\min}$ and $\lambda^{\max}$ are the smallest and the largest eigenvalue of the matrix $A$, respectively.
\end{lemma}

\begin{proof}
Since the matrix $A$ is real symmetric, then all its eigenvalues $\{\lambda_i\}_{i=1}^p$ are real and there exists an orthonormal basis of eigenvectors of $A$ \citep[Theorem 2.1]{treil17} which we denote by $\{v_i\}_{i=1}^p$. We represent $\x=\sum_{i=1}^p \alpha_i v_i$ in such basis where $\alpha_i \in \R$. We use the definition of eigenvalues and the property that the basis is orthonormal:
\begin{align*}
    \x^\intercal\A\x=\left(\sum_{i=1}^p \alpha_i v_i\right)^\intercal A \left(\sum_{i=1}^p \alpha_i v_i\right) = \left(\sum_{i=1}^p \alpha_i v_i\right)^\intercal \left(\sum_{i=1}^p \alpha_i \lambda_iv_i\right) = \sum_{i=1}^p \alpha_i^2 \lambda_i &\le \lambda^{\max}\sum_{i=1}^p \alpha_i^2\\& = \lambda^{\max} \|\x\|_2^2.
\end{align*}
Similarly, we obtain the lower bound: \begin{align*}
    \x^\intercal A\x = \sum_{i=1}^p \alpha_i^2 \lambda_i  \ge \lambda^{\min} \sum_{i=1}^p \alpha_i^2 = \lambda^{\min}\|\x\|_2^2.
\end{align*}

\end{proof}

Next we state a few properties of multivariate normal random vectors. The first lemma describes the distribution of a scalar product between a multivariate normal vector and a constant vector.

\begin{lemma}[Theorem 4.4a, \citet{rencher08}]
\label{lemma:scalar product}
Let $\xi$ denote a multivariate normal random vector in $\R^p$, that is $\xi\sim \N(\mu, \Sigma)$.  For any constant vector $\x\in \R^p$, the scalar product $\xi^\intercal \x$ is a univariate normal random variable with parameters
\begin{align*}
    \xi^\intercal \x \sim \N(\mu^\intercal \x, \x^\intercal \Sigma \x).
\end{align*}
\end{lemma}

The next lemma describes the conditional distribution of two jointly normal random vectors.
\begin{lemma}[Theorem 4.4d, \citet{rencher08}]
\label{lemma:conditional distribution}
    Let $\Theta_1\sim \N(\mu_1, \Sigma_1)$ and $\Theta_2\sim \N(\mu_2,\Sigma_2)$ be two jointly multivariate $p$-dimensional normal random vectors with covariance matrix $\Sigma_{1,2}$. Then, the conditional distribution $\Theta_1 \mid \Theta_2=t$ is a multivariate normal random vector    
    \begin{align*}
        \Theta_1 \mid \Theta_2=t \sim \N\left(\mu_1  + \Sigma_{1,2} \Sigma_{2}^{-1} (t - \mu_2),\,  \Sigma_1 - \Sigma_{1,2}\Sigma_{2}^{-1} \Sigma_{1,2}\right).
    \end{align*}
\end{lemma}

\subsection{Proof of Theorem~\ref{theorem:individual fairness}}
\label{proof:individual fairness}

Before proving Theorem~\ref{theorem:individual fairness}, we provide the expression for the smallest individual fairness constant in \eqref{eq:individual fairness}.
\begin{lemma}
    Let $L^*(\h)$ be the smallest individual fairness constant for the linear model $\h_\theta$ on the set $\calx\subseteq \R^p$.
    \begin{enumerate}
        \item In general, $L^*(\h_\theta) \le \|\theta\|_2$.
        \item If $\calx$ is an open set, then $L^*(\h_\theta) = \|\theta\|_2.$
    \end{enumerate}
\end{lemma}
\begin{proof}
In general, using Cauchy-Schwarz inequality, we can verify that
\begin{align*}
    |\h_\theta(\x) - \h_\theta(\x')| = |\theta^\intercal \x - \theta^\intercal \x'| = |\theta^\intercal (\x - \x')|\le \|\theta\|_2\|\x-\x'\|_2,
\end{align*}
which means that $L^*(\h_\theta)\le \|\theta\|_2$.

Assume now that $\calx\subseteq \R^p$ is an open set. Then for any fixed $\theta\in\R^p$ and for any fixed $x\in\calx$, by assumption that $\calx$ is open, there exists a vector $\x'\in\calx$ in the neighborhood of $\x$ such that $\x - \x'$ is linearly dependent with $\theta$. Therefore, the Cauchy-Schwarz inequality becomes an equality for such pair $\x,\x'\in\calx$, that is
$|\h_\theta(\x) - \h_\theta(\x')| =|\theta^\intercal(\x-\x')|= \|\theta\|_2 \|\x-\x'\|_2$. Hence, $L=\|\theta\|_2$ is the minimal Lipschitz constant of a linear model $\h_\theta$ on the set $\calx$. 
\end{proof}

Now we proof the bound on the norm of private models $\hp$. 
\begin{theorem}

Let $\calm$ be a non-private mechanism returning $\h$ and $\calm^\text{priv}$ be the perturbation mechanism \eqref{eq:general noise model}. Let $\lambda^{\min}_\Sigma$ and $\lambda^{max}_\Sigma$ denote the smallest and the largest eigenvalues of the matrix $\Sigma$ in \eqref{eq:general noise model}.
    \begin{enumerate}
        \item With probability greater than $1-\zeta$ over the randomness of $\calm^\text{priv}$, we have
        \begin{small}
        \begin{align*}
            & \|\theta^\text{priv}\|_2 <  \|\theta\|_2 + \sigma \sqrt{\lambda^{\max}_\Sigma}\sqrt{p + 2\sqrt{p\log\left(\frac 1 \zeta\right)} + 2 \log\left(\frac 1 \zeta\right)}.
        \end{align*}
    \end{small}
        \item With probability greater than $1-\zeta$ over the randomness of $\calm^\text{priv}$, we have
        \begin{small}
        \begin{align*}
            &\|\theta^\text{priv}\|_2 > \max \begin{cases}
                \|\theta\|_2 - \sigma\sqrt{\lambda^{\max}_\Sigma} \sqrt{p + 2\sqrt{p\log\left(\frac 2 \zeta\right)} + 2 \log\left(\frac 2 \zeta\right)},\\
                {\sigma}\sqrt{\lambda^{\min}_\Sigma}\sqrt{\max\left(0, p - 2\sqrt{p\log\left(\frac 2 \zeta\right)}\right)} - \|\theta\|_2.
            \end{cases}
        \end{align*}
        \end{small}
\end{enumerate}
\end{theorem}

\begin{proof}

\textbf{Upper bound.}
We use the triangle inequality to upper bound the norm of the weights of the model $\hp$: 
\begin{align*}
   \|\theta^\text{priv}\|_2= {\|\theta + \sigma \xi\|_2} \le \|\theta\|_2 + {\sigma}\|\xi\|_2 \le \|\theta\|_2+ \sigma \sqrt{\chi^\intercal \Sigma\chi}\le \|\theta\|_2 + \sigma\sqrt{{\lambda^{\max}_\Sigma}}\|\chi\|_2,
\end{align*} 
where $\chi$ is an isotropic Gaussian random variable distributed as $\N(0,\I_p)$. The last inequality is due to Lemma~\ref{lemma:quadratic form}.

We need to bound from above the random variable $\|\chi\|_2$. We use the lower bound on the squared norm of standard Gaussian random vector from \citet[Lemma~1]{laurent00}.  For any $t>0$,
\begin{align*}
    &\Pb_\chi\left(\|\chi\|_2^2 \ge p + 2\sqrt{pt} + 2t\right) \le  \exp(-t)=\zeta,
\end{align*}
which implies that for $t=\log(1/\zeta)$, we have that
\begin{align*}
    \Pb_\chi\left(\|\chi\|_2^2 < p + 2\sqrt{p\log(1/\zeta)}+2\log(1/\zeta)\right) = 1 - \Pb_\chi\left(\|\chi\|_2^2 \ge p + 2\sqrt{p\log(1/\zeta)}+2\log(1/\zeta)\right) > 1-\zeta.
\end{align*}
Hence, with probability greater than $1-\zeta$ over the randomness of $\chi$, we have that
\begin{align*}
    \|\theta^\text{priv}\|_2 < \|\theta\|_2 + \sigma \sqrt{\lambda_\Sigma^{\max}}\sqrt{p + 2\sqrt{p\log(1/\zeta)}+2\log(1/\zeta)}.
\end{align*}

\textbf{Lower bound.} 
We use the reverse triangle inequality to lower bound the individual fairness constant of the private model: \begin{align*}
    \|\theta^\text{priv}\|_2 &= {\|\theta + \sigma \xi\|_2} \ge {\left|\|\theta\|_2 - \sigma\|\xi\|_2\right|}= \max\left(                                                                                                                                                                                                                                          {\sigma}\|\xi\|_2 - \|\theta\|_2,\|\theta\|_2 -  {\sigma}\|\xi\|_2\right)\\
    & =\max\left({\sigma}\sqrt{\chi^\intercal \Sigma \chi} - \|\theta\|_2, \|\theta\|_2 - {\sigma} \sqrt{\chi^\intercal \Sigma \chi}\right)\\
    &  \ge \max\left(\sigma\sqrt{{\lambda^{\min}_{\Sigma}}}\|\chi\|_2 - \|\theta\|_2,\|\theta\|_2 - \sigma\sqrt{{\lambda^{\max}_{\Sigma}}}\|\chi\|_2\right),
\end{align*} 
where $\chi$ is an isotropic Gaussian random variable distributed as $\N(0,\I_p)$.

To complete the proof, we need to bound from below and above the random variable $\|\chi\|_2$. Again, we use the lower and the upper bound on the norm of standard Gaussian random vector from \citet[Lemma~1]{laurent00}.  For any $t>0$,
\begin{align*}
    &\Pb_{\chi}\left(\|\chi\|_2^2 \le p  - 2\sqrt{pt}\right) \le \exp(-t),\\
    &\Pb_\chi\left(\|\chi\|_2^2 \ge p + 2\sqrt{pt} + 2t\right) \le  \exp(-t).
\end{align*}

By combining the lower and the upper bound using the union bound, we finally obtain the two sided bound on $\|\chi\|^2_2$.
\begin{align*}
    &\Pb_\chi\left(\left\{\|\chi\|_2^2 \le p  - 2\sqrt{pt}\right\} \cup \left\{\|\chi\|_2^2 \ge p + 2\sqrt{pt} + 2t\right\}\right)\\
    \cause{union bound}
    &\le \Pb_\chi\left(\|\chi\|_2^2 \le p  - 2\sqrt{pt}\right) + \Pb_\chi\left(\|\chi\|_2^2 \ge p + 2\sqrt{pt} + 2t\right)\\
    &\le \Pb_\chi\left(\|\chi\|_2^2 \le p  - 2\sqrt{pt}\right) + \Pb_\chi\left(\|\chi\|_2^2 \ge p + 2\sqrt{pt} + 2t\right)\\
    & \le 2\exp(-t)=\zeta
\end{align*}
which implies that for $t = \log(2/\zeta)$, we have:                                                  \begin{align*}
    &\Pb_\chi\left(p - 2\sqrt{p \log(2/\zeta)} <  \|\chi\|_2^2 < p + 2\sqrt{p \log(2/\zeta)}+\log(2/\zeta)\right)\\
    &= 1 - \Pb_\chi\left(\left\{\|\chi\|_2^2 \le p  - 2\sqrt{p\log(2/\zeta)  }\right\} \cup \left\{\|\chi\|_2^2 \ge p + 2\sqrt{p\log(2/\zeta)} + 2\log(2/\zeta)\right\}\right)\\
    & > 1-\zeta.
\end{align*}
The rest of the proof consists in taking the squared root over the lower and the upper bound on the random variable $\|\chi\|_2^2$, and substituting it into the lower and the upper bound on $\|\theta^\text{priv}\|_2$.
\end{proof}

\subsection{Proof of Lemma~\ref{lemma:disagreement per data}}

\begin{lemma*}
    The probability that the private models $\hp$ drawn from $\calm^\text{priv}$ disagree with the non-private model $\h$ for a given $\x\in\calx$ is
    \begin{align*}
        \Pb_{\hp}\left[\yh(\hp,\x) \not = \yh(\h,\x)\right] = \Phi\left(\frac{-\left|\amargin(\h,\x,\y)\right|}{\sigma}\right),
    \end{align*}
    where $\Phi$ is the CDF of the standard normal distribution.
\end{lemma*}

\begin{proof}
We rewrite the probability of interest by  using Lemma~\ref{lemma:scalar product} stating that the scalar product of a multivariate normal vector with a constant vector follows a univariate normal distribution :
\begin{align*}
    &\Pb_{\hp}\left(\yh(\hp,\x) \not = \yh(\h,\x)\right) = \Pb_{\hp}\left(\yh(\hp,\x) = 1, \yh(\h,\x)=-1\right) + \Pb_{\hp}\left(\yh(\hp,\x) = -1, \yh(\h,\x)=1\right)\\
    \cause{since the noise $\xi$ in the perturbation mechanism \eqref{eq:general noise model} is independent with the model $\h$}
    &= \Pb_{\hp}\left(\yh(\hp,\x) = 1\right) \cdot \llbracket\yh(\h,\x)=-1\rrbracket + \Pb_{\hp}\left(\yh(\hp,\x) = -1\right) \cdot \llbracket\yh(\h,\x)=1\rrbracket\\
    \cause{the brackets $\llbracket\cdot\rrbracket$ denote the indicator function}
    & = \mathop\Pb_{\xi\sim\N(0,\Sigma)}\left(\theta^\intercal \x + \sigma \xi^\intercal \x \ge 0\right) \llbracket\theta^\intercal \x < 0\rrbracket + \mathop\Pb_{\xi\sim\N(0,\Sigma)}\left(\theta^\intercal \x + \sigma \xi^\intercal \x < 0\right) \llbracket\theta^\intercal \x \ge 0\rrbracket\\
    & = \left(1 - \Phi\left(\frac{-\theta^\intercal \x}{\sigma\sqrt{\x^\intercal \Sigma \x}}\right)\right)\llbracket\theta^\intercal \x < 0\rrbracket + \Phi\left(\frac{-\theta^\intercal \x}{\sigma\sqrt{\x^\intercal \Sigma \x}}\right) \llbracket\theta^\intercal \x \ge 0\rrbracket \\
    \cause{using the property that $1-\Phi(w) = \Phi(-w)$} 
    & = \Phi\left(\frac{\theta^\intercal \x}{\sigma\sqrt{\x^\intercal \Sigma \x}}\right) \llbracket\theta^\intercal \x < 0\rrbracket + \Phi\left(\frac{-\theta^\intercal \x}{\sigma\sqrt{\x^\intercal \Sigma \x}}\right)\llbracket\theta^\intercal \x \ge 0\rrbracket = \Phi\left(\frac{-|\theta^\intercal \x|}{\sigma\sqrt{\x^\intercal \Sigma \x}}\right)\\
    &= \Phi\left(\frac{-|\amargin(\h,\x,\y)|}{\sigma}\right).
\end{align*}
In the last equality, we use the notation for angular margins $\amargin(\h,\x,y) = \frac{\y \theta^\intercal\x}{\sqrt{\x^\intercal\Sigma\x}}$ as it simplifies the exposition and is useful to present further results of the paper. We put the label $\y$ in the definition of $\alpha$ since  $\y\in\caly=\{-1,1\}$.
\end{proof}

\subsection{Proof of Theorem~\ref{theorem:disagreement number}}

\begin{theorem*}[Disagreement ratio bound]
    With probability greater than $1-\zeta$ over the randomness of  $\calm^\text{priv}$, the disagreement ratio of $\hp$ is bounded
    \begin{align*}
        \Pb_\cald\left[\yh(\hp,\X) \not = \yh(\h,\X)\right] <\frac{\E_\cald\left[\Phi\left(\frac{-\left|\amargin(\h,\X,\Y)\right|}{\sigma}\right)\right]}{\zeta}.
    \end{align*}
\end{theorem*}

\begin{proof}
We first use Markov's inequality to obtain the bound on the tail of the disagreement ratio:
\begin{align*}
    \Pb_{\hp}\left[\Pb_\cald [\yh(\hp,\X) \not = \yh(\h,\X)] \ge t\right] \le \frac{\E_{\hp}\left[\Pb_\cald [\yh(\hp,\X) \not = \yh(\h,\X)]\right]}{t} = \zeta.
\end{align*}
Hence, with probability greater than $1-\zeta$ over the randomness of the perturbation mechanism $\calm^\text{priv}$, we have
\begin{align*}
    \Pb_\cald [\yh(\hp,\X) \not = \yh(\h,\X)] < \frac{\E_{\hp}\left[\Pb_\cald [\yh(\hp,\X) \not = \yh(\h,\X)]\right]}{\zeta}.
\end{align*}
To derive the expression on the right hand side, we use the Fubini-Tonelli's theorem to change the order of expectation:
\begin{align*}
    \E_{\hp}\left[\Pb_\cald [\yh(\hp,\X) \not = \yh(\h,\X)]\right] &=     \E_{\hp}\left[\E_\cald \llbracket\yh(\hp,\X) \not = \yh(\h,\X)\rrbracket\right] \\&= \E_\cald \left[\E_{\hp}\llbracket\yh(\hp,\X) \not = \yh(\h,\X)\rrbracket\right]\\
    &= \E_{\cald}\left[\Phi\left(\frac{-|\amargin(\h,\X,\Y)|}{\sigma}\right)\right],
\end{align*}
where the last equality follows from Lemma~\ref{lemma:disagreement per data}.

\end{proof}

\subsection{Proof of Lemma~\ref{lemma:first moment}}
\label{proof:first moment}

\begin{lemma}[Expected fairness]
    \label{lemma:expectation general}
    Let $\calm$ be a non-private mechanism returning $\h$ and $\calm^\text{priv}$ be the perturbation mechanism \eqref{eq:general noise model}.
    \begin{enumerate}
        \item The expected accuracy $\acc$ over the randomness of private models $\hp$ equals
          \begin{align*}
              \cale_{\acc}(\h,\sigma,\Sigma,\cald) = \E_{\cald}\left[\Phi\left(\frac{\amargin(\h,\X,\Y)}{\sigma}\right)\right].
          \end{align*}
        \item The expected fairness $\fk$ over the randomness of private models $\hp$ equals
        \begin{align*}
            \cale_{\fk}(\h,\sigma,\Sigma,\cald)  =C_k^0 + \sum_{k'=1}^K C_k^{k'}\cale_{\acc}(\h,\sigma,\Sigma,\cald_{k'}).\nonumber
        \end{align*}
    \end{enumerate}
\end{lemma}

Before proving Lemma~\ref{lemma:first moment}, we need to prove the following technical lemma. 
\begin{lemma}\label{lemma:probability correct prediction}
Let $\calm$ be a non-private mechanism returning $\h$ and $\calm^\text{priv}$ be the perturbation mechanism \eqref{eq:general noise model} which generates private models $\hp$. The probability that the private models  predict the label $\y$ given an example $\x$ is equal to
\begin{align*}
    \Pb_{\hp}\left(\yh(\hp,\x)=\y\right) = \Phi\left(\frac{\amargin(\h,\x,\y)}{\sigma}\right).
\end{align*}
\end{lemma}

\begin{proof}
    We rewrite the probability that the private model makes a prediction  $\y$ for a given example $\x\in\calx$:
\begin{align*}
    &\Pb_{\hp}\left(\yh(\hp,\x)=\y\right) = \Pb_{\hp}\left(\yh(\hp,\x)=1, \y=1\right) + \Pb_{\hp}\left(\yh(\hp,\x)=-1,\y=-1\right)\\
    \cause{since the noise $\xi$ in perturbation mechanism \eqref{eq:general noise model} is independent with the label $\y$}
    &= \Pb_{\hp}\left(\yh(\hp,\x)=1\right)\llbracket \y=1\rrbracket + \Pb_{\hp}\left(\yh(\hp,\x)=-1\right)\llbracket \y=-1\rrbracket \\
    \cause{the brackets $\llbracket\cdot\rrbracket$ denote the indicator function}
    & = \mathop{\Pb}_{\xi\sim\N(0,\Sigma)}\left(\theta^\intercal \x + \sigma \xi^\intercal \x \ge 0 \right)\llbracket \y=1\rrbracket + \mathop{\Pb}_{\xi\sim\N(0,\Sigma)}\left(\theta^\intercal \x + \sigma \xi^\intercal \x < 0 \right)\llbracket \y=-1\rrbracket\\
    \cause{using Lemma~\ref{lemma:scalar product}}
    & = \left(1 - \Phi\left(\frac{-\theta^\intercal \x}{\sigma\sqrt{\x^\intercal \Sigma\x}}\right)\right)\llbracket \y=1 \rrbracket + \Phi\left(\frac{-\theta^\intercal \x}{\sigma\sqrt{\x^\intercal \Sigma\x}}\right) \llbracket\y=-1\rrbracket\\
    \cause{using the property that $1-\Phi(w) = \Phi(-w)$ and the definition of angular margins $\amargin(\h,\x,\y)$}
    & = \Phi\left(\frac{\theta^\intercal \x}{\sigma\sqrt{\x^\intercal \Sigma\x}}\right)\llbracket \y=1\rrbracket + \Phi\left(\frac{-\theta^\intercal \x}{\sigma\sqrt{\x^\intercal \Sigma\x}}\right) \llbracket \y=-1\rrbracket = \Phi\left(\frac{\y\theta^\intercal \x}{\sigma\sqrt{\x^\intercal \Sigma\x}}\right) = \Phi\left(\frac{\amargin(\h,\x,\y)}{\sigma}\right).
\end{align*}
\end{proof}

Now we are ready to prove Lemma~\ref{lemma:expectation general}.
\begin{proof}[Proof of Lemma~\ref{lemma:expectation general}]
The proof of the first part consists in changing the order of expectation using Fubini-Tonelli's theorem:
\begin{align*}
     \cale_{\acc} = \E_{\hp} \left[ \acc (\hp, \cald)\right] 
     &=  \E_{\hp}\left(\E_{\cald}\llbracket\yh(\hp,\X) = \Y\rrbracket \right)  
     = \E_{\cald}\left(\E_{\hp}\llbracket\yh(\hp,\X)=\Y\rrbracket\right)\\
     & = \E_{\cald}\left[\Pb_{\hp}\left(\yh(\hp,\X)=\Y\right)\right]= \E_{\cald}\left[\Phi\left(\frac{\amargin(\h,\X,\Y)}{\sigma}\right)\right],
\end{align*}
where the last equality is due to Lemma~\ref{lemma:probability correct prediction}.

To prove the second part of the lemma, we use  the definition of fairness measure $\fk(\h,\cald)= C_k^0 + \sum_{k'=1}^K C_{k}^{k'} \cdot \acc(\h,\cald_{k'})$, the obtained expression for the expected accuracy $\cale_{\acc}$, and the linearity of expectation.
\end{proof}

\subsection{Proof of Lemma~\ref{lemma:variance of fairness bound}}
\label{proof:variance of fairness bound}
\begin{lemma}[Variance of fairness of private models $\hp$]
    \label{lemma:variance general}
    Let $\calm$ be a non-private mechanism returning $\h$ and $\calm^\text{priv}$ be the perturbation mechanism \eqref{eq:general noise model}.
    \begin{enumerate}
        \item The variance of the accuracy $\acc$ over the randomness of the privacy-preserving mechanism $\calm^\text{priv}$ is bounded from above, that is
        % \begin{small}
            \begin{align*}
             &\V_{\hp}\left[\acc(\hp,\cald)\right]\le \calv_{\acc}(\h,\sigma, \Sigma,\cald) = {\mathop{\E}_{\Z^{(1)},\Z^{(2)}\sim \cald}\left[\Phi\left(\frac{\amargin(\h,\Z^{\text{min}}}{\sigma}\right) \Phi\left(\frac{-\amargin(\h,\Z^{\text{max}})}{\sigma}\right)\right]},
            \end{align*}
        % \end{small}
        where 
        $\Z^{\text{min}} = \argmin_{\Z^{(1)}, \Z^{(2)}} \amargin(\h,\Z)$ and  $\Z^{\text{max}} = \argmax_{\Z^{(1)}, \Z^{(2)}} \amargin(\h,\Z)$.
    \item The variance of the group fairness measure $\calf_k$ over the randomness of the privacy-preserving mechanism $\calm^\text{priv}$ is bounded from above, that is
    % \begin{small}
        \begin{align*}
            &\V_{\hp}\left[\fk(\hp,\cald)\right] \le \calv_{\fk}(\h,\sigma,\Sigma, \cald)=\left[\sum_{k'=1}^K |C_{k}^{k'}| \sqrt{\calv_{\acc}(\h,\sigma,\Sigma, \cald_{k'})}\right]^2.
        \end{align*}
    % \end{small}
    \end{enumerate}
\end{lemma}

\begin{proof}
We start with the derivation of the accuracy variance bound.

\textbf{Accuracy variance.}
Using the definition of accuracy,
\begin{align*}
    \Var_{\hp}\left(\acc(\hp,\cald_k)\right) = \Var_{\hp}\left(\E_{\cald_k}\llbracket \yh(\hp,\X)=\Y\rrbracket\right) &= \E_{\hp}\left(\left(\E_{\cald_k}\llbracket \yh(\hp,\X)=\Y\rrbracket\right)^2\right)\\
    &- \left(\E_{\hp}\left(\E_{\cald_k}\llbracket \yh(\hp,\X)=\Y\rrbracket\right)\right)^2. 
\end{align*}
We would like to change the order of expectation from the expectation over $\hp$ to the expectation over data distribution $\cald$. To be able to do so, we introduce two independent identically distributed random variables $\Z'\sim\cald_k$ and $\Z''\sim\cald_k$. We upper bound the first term in the expression for the accuracy variance:
\begin{align*}
    &\E_{\hp}\left(\mathop{\E}_{\Z'\sim\cald_k}\llbracket \yh(\hp,\X')=\Y'\rrbracket \cdot  \mathop{\E}_{\Z''\sim\cald_k}\llbracket \yh(\hp,\X'')=\Y''\rrbracket\right)\\
    \cause{independence of $\Z'$ and $\Z''$}
    &=  \E_{\hp}\left(\mathop{\E}_{\Z'\sim\cald_k,\Z''\sim\cald_k}\llbracket \yh(\hp,\X')=\Y', \yh(\hp,\X'')=\Y''\rrbracket\right)\\
    \cause{changing the order of expectation using  Fubini-Tonelli's theorem}
    & = \mathop{\E}_{\Z'\sim\cald_k,\Z''\sim\cald_k}\left( \E_{\hp} \llbracket \yh(\hp,\X')=\Y', \yh(\hp,\X'')=\Y''\rrbracket\right) \\&= \mathop{\E}_{\Z'\sim\cald_k,\Z''\sim\cald_k}\left( \Pb_{\hp} \left( \yh(\hp,\X')=\Y', \yh(\hp,\X'')=\Y''\right)\right)\\
    \cause{using the property that $\Pb(A,B)\le \min(\Pb(A), \Pb(B))$}
    &\le \mathop{\E}_{\Z'\sim\cald_k,\Z''\sim\cald_k}\left( \min \left(\Pb_{\hp}\left[\yh(\hp,\X')=\Y'\right], \Pb_{\hp}\left[\yh(\hp,\X'')=\Y''\right]\right)\right)\\
    \cause{using Lemma~\ref{lemma:probability correct prediction}}
    & = \mathop{\E}_{\Z'\sim\cald_k,\Z''\sim\cald_k}\left( \min \left(\Phi\left(\frac{\amargin(\h,\X',\Y')}{\sigma}\right), \Phi\left(\frac{\amargin(\h,\X'',\Y'')}{\sigma}\right)\right)\right).
\end{align*}
We rewrite the second term in the expression for the accuracy variance:
\begin{align*}
&\E_{\hp}\left(\E_{\Z'\sim\cald_{k}}\llbracket\yh(\hp,\X')=\Y'\rrbracket\right)\cdot \E_{\hp}\left(\E_{\Z''\sim\cald_{k}}\llbracket\yh(\hp,\X'')=\Y''\rrbracket\right)=\\
\cause{changing the order of expectations using Fubini-Tonelli's theorem}
& =\E_{\Z'\sim\cald_{k}}\left(\E_{\hp}\llbracket\yh(\hp,\X')=\Y'\rrbracket\right)\cdot \E_{\Z''\sim\cald_{k}}\left(\E_{\hp}\llbracket\yh(\hp,\X'')=\Y''\rrbracket\right)\\
\cause{using Lemma~\ref{lemma:probability correct prediction}}
&=\E_{\Z'\sim\cald_{k}}\left(\Phi\left(\frac{\amargin(\h,\X',\Y')}{\sigma}\right)\right)\cdot \E_{\Z''\sim\cald_{k}}\left(\Phi\left(\frac{\amargin(\h,\X'',\Y'')}{\sigma}\right)\right)\\
\cause{independence of $\Z'$ and $\Z''$}
& = \mathop{\E}_{\Z'\sim\cald_k,\Z''\sim\cald_{k}}\left(\Phi\left(\frac{\amargin(\h,\X',\Y')}{\sigma}\right)\cdot \Phi\left(\frac{\amargin(\h,\X'',\Y'')}{\sigma}\right)\right).
\end{align*}

Finally, by introducing the notation
$\displaystyle \Z^{\text{min}} = \argmin_{\Z', \Z''} \amargin(\h,\Z)$,  $\displaystyle \Z^{\text{max}} = \argmax_{\Z', \Z''} \amargin(\h,\Z)$, we show the following upper bound on the variance of accuracy:
\begin{align*}
    \Var_{\hp}\left[\acc(\hp,\cald_k)\right] \le&\mathop{\E}_{\Z'\sim\cald_k,\Z''\sim\cald_k}\left(\Phi\left(\frac{\amargin(\h,\X^\text{min},\Y^\text{min})}{\sigma}\right) -  \Phi\left(\frac{\amargin(\h,\X^\text{min},\Y^\text{min})}{\sigma}\right)\cdot \Phi\left(\frac{\amargin(\h,\X^\text{max},\Y^\text{max})}{\sigma}\right)\right)\\
    & = \mathop{\E}_{\Z'\sim\cald_k,\Z''\sim\cald_k}\left(\Phi\left(\frac{\amargin(\h,\X^\text{min},\Y^\text{min})}{\sigma}\right) \cdot \left( 1 - \Phi\left(\frac{\amargin(\h,\X^\text{max},\Y^\text{max})}{\sigma}\right)\right)\right)\\
    \cause{using the property that $1-\Phi(w) = \Phi(-w)$}
    & = \mathop{\E}_{\Z'\sim\cald_k,\Z''\sim\cald_k}\left(\Phi\left(\frac{\amargin(\h,\Y^\text{min},\Y^\text{min})}{\sigma}\right) \cdot \Phi\left(\frac{-\amargin(\h,\X^\text{max},\Y^\text{max})}{\sigma}\right)\right) \\&= \calv_{\acc}(\h,\sigma,\Sigma,\cald_k).
\end{align*}

\textbf{Fairness variance.}
We use the distributive property of the covariance to derive the fairness variance:
\begin{align*}
    \Var_{\hp}\left[\calf_k(\hp,\cald)\right] &= \Cov_{\hp}\left[\calf_k(\hp,\cald), \calf_k(\hp,\cald)\right] \\&= \sum_{k'=1}^K\sum_{k''=1}^K C_k^{k'}C_{k}^{k''}\Cov_{\hp}\left[\acc(\hp,\cald_{k'}),\acc(\hp,\cald_{k''})\right].
\end{align*}
 After, we use the triangle inequality and the covariance inequality to upper bound the fairness variance:
\begin{align*}
    \Var_{\hp}\left(\calf_k(\hp,\cald)\right) &\le \sum_{k'=1}^K\sum_{k''=1}^K |C_k^{k'}|\cdot|C_{k}^{k''}|\cdot\left|\Cov_{\hp}\left[\acc(\hp,\cald_{k'}),\acc(\hp,\cald_{k''})\right]\right|\\
    \cause{using the covariance inequality}
    &\le \sum_{k'=1}^K\sum_{k''=1}^K |C_k^{k'}|\cdot|C_{k}^{k''}|\cdot\left|\sqrt{\Var_{\hp}\left[\acc(\hp,\cald_{k'})\right]\cdot\Var_{\hp}\left[\acc(\hp,\cald_{k''})\right]}\right|\\
    & = \left(\sum_{k'=1}^K|C_k^{k'}|\sqrt{\Var_{\hp}\left[\acc(\hp,\cald_{k'})\right]} \right)^2  \le \left(\sum_{k'=1}^K|C_k^{k'}|\sqrt{\calv_{\acc}(\h,\sigma,\Sigma,\cald_{k'})} \right)^2 \\&= \calv_{\fk}(\h,\sigma,\Sigma,
    \cald).
\end{align*}
The last inequality is due to the upper bound on the accuracy variance proven in the first part of the lemma.

\end{proof}

\subsection{Proof of Theorem~\ref{theorem:confidence bounds}}
\label{proof:confidence bounds}

We restate Theorem~\ref{theorem:confidence bounds} by additionally providing the high probability bound on accuracy of private models.

\begin{theorem*}Let $\calm$ be a non-private mechanism returning $\h$ and $\calm^\text{priv}$ be the perturbation mechanism \eqref{eq:general noise model}.
    \begin{enumerate}
    \item Let $\cale_{\fk}$ and $\calv_{\fk}$ denote the expected fairness and the fairness variance  upper bound defined in Lemma~\ref{lemma:expectation general} and  Lemma~\ref{lemma:variance general}. With probability at least $1-\zeta$ over the randomness of $\calm^\text{priv}$, we have
    \begin{align*}
        \begin{cases}
            \fk(\hp,\cald)  <  \cale_{\fk}(\h,\sigma,\Sigma,\cald) + \sqrt{\frac{\calv_{\fk}(\h,\sigma,\Sigma,\cald)}{\zeta}}, \\
            \fk(\hp,\cald)  > \cale_{\fk}(\h,\sigma,\Sigma,\cald) - \sqrt{\frac{\calv_{\fk}(\h,\sigma,\Sigma,\cald)}{\zeta}}.
        \end{cases}
    \end{align*}
    \item Let $\cale_{\acc}$ and $\calv_{\acc}$ denote the expected fairness and the fairness variance  upper bound defined in Lemma~\ref{lemma:expectation general} and  Lemma~\ref{lemma:variance general}. With probability at least $1-\zeta$ over the randomness of $\calm^\text{priv}$, we have
    \begin{align*}
        \begin{cases}
            \acc(\hp,\cald)  <  \cale_{\acc}(\h,\sigma,\Sigma,\cald) + \sqrt{\frac{\calv_{\acc}(\h,\sigma,\Sigma,\cald)}{\zeta}}, \\
            \acc(\hp,\cald)   > \cale_{\acc}(\h,\sigma,\Sigma,\cald) - \sqrt{\frac{\calv_{\acc}(\h,\sigma,\Sigma,\cald)}{\zeta}}.
        \end{cases}
    \end{align*}
    \end{enumerate}
\end{theorem*}

\begin{proof}
We use the Chebyshev's inequality to bound the fairness and the accuracy of private models. For any $t > 0$, we have:
\begin{align*}
    \Pb_{\hp} \left(|\calf_k(\hp,\cald) - \E_{\hp}\left(\calf_k(\hp,\cald)\right)|\ge t \right) \le \Var_{\hp}\left(\calf_k(\hp,\cald)\right) / t^2 \le \calv_{\fk}(\h,\sigma,\Sigma,
    \cald)/t^2= \zeta.
\end{align*}
Hence, with probability greater than $1-\zeta$ over the randomness of the perturbation mechanism \eqref{eq:general noise model}, we have:
\begin{align*}
    |\calf_k(\hp,\cald) - \cale_{\fk}(\h,\sigma,\Sigma,
    \cald)| < \sqrt{\calv_{\fk}(\h,\sigma,\Sigma,
    \cald) / \zeta},
\end{align*}
where above we use the notation for the expected fairness $\cale_{\fk}$ from Lemma~\ref{lemma:expectation general}.

Using the same proof technique, we derive the high probability bound on the accuracy of private models. That is, with probability greater than $1-\zeta$, we have
\begin{align*}
    |\acc(\hp,\cald) - \cale_{\acc}(\h,\sigma,\Sigma,
    \cald)| < \sqrt{\calv_{\acc}(\h,\sigma,\Sigma,
    \cald) / \zeta}.
\end{align*}
\end{proof}

\subsection{Proof of Lemma~\ref{theorem:bayesian estimate}}
\label{proof:bayesian estimate}

We prove a more general result that in the main text. We consider the prior on $\theta$ with a mean $\mu$ and a non-diagonal symmetric positive-definite covariance matrix $\eta^2A$, that is $\Theta \sim \N(\mu, \eta^2 A)$. 
\begin{theorem*}[Bayesian estimation of the private model]
   Let $\hp$ be the private model  obtained from the model $\h$ using the perturbation mechanism \eqref{eq:general noise model}. Assume a Bayesian prior on model weights  $\Theta \sim \N(\mu, \eta^2A)$. Then, the posterior distribution of model weights $\h$ follows a multivariate Gaussian distribution 
    \begin{align*}
        \Theta \mid \Theta^\text{priv}=\theta^\text{priv}, \sigma,\Sigma \sim \N\left(\mu+A\left[A+\frac{\sigma^2}{\eta^2}\Sigma\right]^{-1}(\theta^\text{priv}-\mu),A\left[A + \frac{\sigma^2}{\eta^2}\Sigma \right]^{-1}\sigma^2\Sigma\right).
    \end{align*}
    In particular, for the uniform prior, that is $\eta^2 \to \infty$, we have $\Theta \mid \Theta^\text{priv}=\theta^\text{priv}, \sigma, \Sigma \sim \N\left(\theta^\text{priv},\sigma^2\Sigma\right)$.
\end{theorem*}

\begin{proof}

Due to our assumption that $\Theta \sim \N(\mu, \eta^2A)$, the marginal distribution of the private model weights $\Theta^\text{priv}$ is a multivariate normal Gaussian vector:
\begin{align*}
    \Theta^\text{priv}=\Theta + \sigma \xi  \sim \N\left(\mu, \eta^2 A + \sigma^2 \Sigma \right).
\end{align*}
Since the noise vector $\xi$ is independent with $\Theta$, the covariance matrix for the random vectors  $\Theta$ and  $\Theta^\text{priv}$ is
\begin{align*}
\Cov\left[\Theta, \Theta^\text{priv}\right] = \eta^2 A.  
\end{align*}
Next, we use Lemma~\ref{lemma:conditional distribution} which describes the parameters of the conditional distribution of two correlated multivariate normal random vectors:
\begin{align*}
    \E\left[\Theta \mid \Theta^\text{priv}=\theta^\text{priv}\right] &= \mu + \eta^2A (\eta^2 A + \sigma^2\Sigma)^{-1}\left(\theta^\text{priv}-\mu\right)=\mu+A\left(A +\frac{\sigma^2}{\eta^2}\Sigma\right)^{-1}(\theta^\text{priv}-\mu),\\
    \Cov[\Theta, \Theta \mid \Theta^\text{priv}=\theta^\text{priv}] &= \eta^2 A - \eta^2 A \left[\eta^2A + \sigma^2\Sigma\right]^{-1} \eta^2 A = \eta^2 A \left( \I_p - \left[\eta^2A + \sigma^2\Sigma\right]^{-1} \eta^2 A\right)\\
    &= \eta^2 A \left[\eta^2A + \sigma^2\Sigma\right]^{-1} \left(\eta^2 A + \sigma^2\Sigma - \eta^2 A \right)= A \left[A + \frac{\sigma^2}{\eta^2} \Sigma\right]^{-1}\sigma^2 \Sigma.
\end{align*}
Finally, we note that when $\eta^2\to\infty$, which corresponds to the case of the uniform prior assumption, we have:
\begin{align*}
    &\E[\Theta\mid\Theta^\text{priv}=\theta^\text{priv}] \xrightarrow{\eta\to\infty}\mu + AA^{-1}(\theta^\text{priv} - \mu) = \theta^\text{priv},\\ 
    &\Cov[\Theta, \Theta \mid \Theta=\theta^\text{priv}] \xrightarrow{\eta\to\infty} A A^{-1}\sigma^2 \Sigma = \sigma^2 \Sigma.
\end{align*}
Hence, $\Theta \mid \Theta^\text{priv}=\theta^\text{priv} \sim \N\left(\theta^\text{priv}, \sigma^2\Sigma\right).$
\end{proof}

\subsection{Proof of Lemma~\ref{lemma:noisy sgd}}

We restate the assumptions of Section~\ref{ssec:noisy sgd}

\paragraph{(A1)} The empirical loss is well-approximated by a quadratic function, that is $\mathcal{\widehat L}(\theta) = \mathcal{\widehat  L}(\theta^*) + \frac 1 2 (\theta-\theta^*)^\intercal H (\theta-\theta^*),$
where $H_{ij}=\frac{\partial^2_\theta \mathcal{\widehat L}(\theta)}{\partial\theta_i\partial\theta_j}|_{\theta=\theta^*}$ is the symmetric positive definite Hessian at the optimum $\theta^* = \arg\min_{\theta\in\R^p}\mathcal{\widehat L}(\theta)$. 
\paragraph{(A2)} The noisy GD dynamics is well-approximated by its continuous-time dynamics
\begin{align}
\label{eq:OL process}
    d\theta_t = -\eta \left(H (\theta_t-\theta^*)dt +  \sigma \I_p dW_t\right),
\end{align}
where $dW_t$ is the Wiener process. 

\begin{lemma}
    % \label{lemma:noisy sgd}
    Under assumptions (A1)--(A2), the stationary distribution of the model weights of noisy GD mechanism is multivariate Gaussian, that is
        $\theta^\text{priv} \sim \N\left(\theta^*,\frac 1 2 {\sigma^2\eta}   H^{-1}\right).$
\end{lemma}

\begin{proof}
    
Under assumptions (A1)--(A2), it is shown that the stochastic process \eqref{eq:OL process} has an analytic stationary solution which is a  multivariate normal distribution $\theta \sim \N(\theta^*, \Theta)$, 
where the covariance matrix $\Theta$ can be found from the condition \citep{godreche19}:
\begin{align}
   \label{eq:lyapunov}
   \Theta H + H \Theta = \eta  \sigma^2 \I_p.
\end{align}
While in general the equation cannot be solved analytically, for our case, however, such analytical solution exists. We can verify by substitution that  $\Theta = \frac  1 2 {\eta\sigma^2} H^{-1}$ which concludes the proof.
\end{proof}

\subsection{Proof of Lemma~\ref{theorem:finite sample}}

\begin{lemma}[Corollary of Lemma~3.4 from \citep{mangold22}]
    % \label{theorem:finite sample}
    Let $\hp$ be the private model  obtained from the model $\h$ using the  perturbation mechanism \eqref{eq:general noise model}.
    Assume that $n \ge \frac{8\log((2K+1) / \kappa)}{\min_{k'} p_{k'}}$ where $p_{k'}$ is the true proportion of examples from group $k'$. Assume also that $\Pb_{D}\left(\sum_{k'=0}^K\left|C_k^{k'} - \widehat C_{k}^{k'}\right|>\alpha_C\right)\le B_3 \exp(-B_4 \alpha_C^2n)$.
    With probability $1-\kappa$ over the randomness of dataset $D$ of size $n$, with probability $1-\zeta$ over the randomness of private models $\hp$,  we have 
    \begin{align*}
        \left|\fk(\hp,\cald) - \cale_{\fk}(\h,\sigma,D)\right| \le \sqrt{{\calv_{\fk}(\h,\sigma,D)}/{\zeta}} + O\left(\sum_{k'=1}^K |\widehat C_k^{k'}|\sqrt{\frac{d_{\calh} + \log(K/\kappa)}{n p_{k'}}}\right),
    \end{align*}
    where $d_{\calh}$ is the Natarajan dimension of the class of linear models $\calh$.
\end{lemma}

\begin{proof}

For any given dataset $D$, we prove in Theorem~\ref{theorem:confidence bounds} that
\begin{align*}
    \Pb_{\hp}\left(\cale_{\fk}(\h,\sigma,D) - \sqrt{\frac{\calv_{\fk}(\h,\sigma,D)}{\zeta}}  \le \fk(\hp) \le \cale_{\fk}(\h,\sigma,D) + \sqrt{\frac{\calv_{\fk}(\h,\sigma,D)}{\zeta}}\right) \ge 1- \zeta.
\end{align*}
It implies that with probability $1$ over the randomness of dataset $D$, the above inequality holds.

To simplify the notation, further in the proof, we denote fairness on dataset as $\widehat \fk(\h) = \fk(\h, D)$ and we denote fairness on distribution $\cald$ as $\fk(\h) = \fk(\h,\cald)$. We also use the notation $\cale_{\widehat\fk}(\h)=\cale_{\fk}(\h,\sigma,\D)$ and $\calv_{\widehat\fk}(\h)=\calv_{\fk}(\h,\sigma,\D)$ to denote expectation and variance of fairness on the dataset $D$.

Next, we bound the error in the estimation of exact fairness by its empirical counterpart $\widehat \fk$, that is 
\begin{align}
    \label{eq:sup bound}
    \Pb_{D}\left(\left|\widehat \fk(\hp) - \fk(\hp)\right|\ge t \right) \le  \Pb_{D}\left(\sup_{\h\in\calh}\left|\widehat \fk(\h) - \fk(\h)\right|\ge t \right) \le \kappa.
\end{align}
We use the proof of \citep[Appendix D, proof of Lemma~3.4]{mangold22} who show  that for a fixed $\kappa$, for $n \ge \frac{8\log\left(\frac{2K+1}{\kappa}\right)}{\min_{k'\in\{1,\dots,K\}}p_{k'}}$, and under assumption that $\Pb_{D}\left(\sum_{k'=0}^K\left|C_k^{k'} - \widehat C_{k}^{k'}\right|>\alpha_C\right)\le B_3 \exp(-B_4 \alpha_C^2n)$, inequality \eqref{eq:sup bound} holds for
\begin{align*}
    t = \sqrt{\frac{\log\left(\frac{B_3(2K+1)}{\kappa}\right)}{B_4n}} + \sum_{k'=1}^K 8|\widehat C_{k}^{k'}| \sqrt{\frac{d_{\calh}\log\left(\frac{np_{k'}}{2} + 2\log(|\caly|)\right) + \log\left(\frac{8(2K+1)}{\kappa} \right)}{np_{k'}}},
\end{align*}
where $d_{\calh}$ is the Natarajan dimension of the class $\calh$.

Consider then the following probability $P$ of the intersection of two events 
\begin{small}
\begin{align*}
    P &=\Pb_{D}\left(\left\{\Pb_{\hp}\left(\cale_{\widehat\fk}(\h) - \sqrt{\frac{\calv_{\widehat\fk}(\h)}{\zeta}}  \le \widehat\fk(\hp) \le \cale_{\widehat\fk}(\h) + \sqrt{\frac{\calv_{\widehat\fk}(\h)}{\zeta}}\right) \ge 1- \zeta\right\}\cap\left\{\sup_{\h\in\calh}\left|\widehat \fk(\h) - \fk(\h)\right|\le t \right\}\right)\\
    & = 1 - \Pb_{D}\left(\left\{\Pb_{\hp}\left(\cale_{\widehat\fk}(\h) - \sqrt{\frac{\calv_{\widehat\fk}(\h)}{\zeta}}  \le \widehat\fk(\hp) \le \cale_{\widehat\fk}(\h) + \sqrt{\frac{\calv_{\widehat\fk}(\h)}{\zeta}}\right) < 1- \zeta\right\}\cup\left\{\sup_{\h\in\calh}\left|\widehat \fk(\h) - \fk(\h)\right| \ge t \right\}\right)\\
    \cause{union bound}
    & \ge 1 - \Pb_{D}\left(\Pb_{\hp}\left(\cale_{\widehat\fk}(\h) - \sqrt{\frac{\calv_{\widehat\fk}(\h)}{\zeta}}  \le \widehat\fk(\hp) \le \cale_{\widehat\fk}(\h) + \sqrt{\frac{\calv_{\widehat\fk}(\h)}{\zeta}}\right) < 1- \zeta\right) - \Pb_D\left(\sup_{\h\in\calh}\left|\widehat \fk(\h) - \fk(\h)\right| \ge t \right)\\
    &\ge 1 -\kappa.
\end{align*}
\end{small}

The rest of the proof consists in observing that for the probability $P$, we have the following upper bound which represents our probability of interest
\begin{small}
\begin{align*}
    P&=\Pb_{D}\left(\left\{\Pb_{\hp}\left(\cale_{\widehat\fk}(\h) - \sqrt{\frac{\calv_{\widehat\fk}(\h)}{\zeta}}  \le \widehat\fk(\hp) \le \cale_{\widehat\fk}(\h) + \sqrt{\frac{\calv_{\widehat\fk}(\h)}{\zeta}}\right) \ge 1- \zeta\right\}\cap\left\{\sup_{\h\in\calh}\left|\widehat \fk(\h) - \fk(\h)\right|\le t \right\}\right)\\
    \cause{adding $\fk(\hp) - \widehat\fk(\hp)$ in the first inequality, and using the property that $\sup_{\h\in\calh}\left|\widehat \fk(\h) - \fk(\h)\right|\le t$}
    & \le  \Pb_{D}\left(\left\{\Pb_{\hp}\left(\cale_{\widehat\fk}(\h) - \sqrt{\frac{\calv_{\widehat\fk}(\h)}{\zeta}} -t  \le \fk(\hp) \le \cale_{\widehat\fk}(\h) + \sqrt{\frac{\calv_{\widehat\fk}(\h)}{\zeta}} + t\right) \ge 1- \zeta\right\}\cap\left\{\sup_{\h\in\calh}\left|\widehat \fk(\h) - \fk(\h)\right|\le t \right\}\right)\\
    \cause{using $\Pb(A\cap B) \le \Pb(A)$}
    &\le \Pb_{D}\left(\Pb_{\hp}\left(\cale_{\widehat\fk}(\h) - \sqrt{\frac{\calv_{\widehat\fk}(\h)}{\zeta}} -t  \le \fk(\hp) \le \cale_{\widehat\fk}(\h) + \sqrt{\frac{\calv_{\widehat\fk}(\h)}{\zeta}} + t\right) \ge 1- \zeta\right).
\end{align*}
\end{small}
Hence, we show that
\begin{align*}
    \Pb_{D}\left(\Pb_{\hp}\left(\cale_{\widehat\fk}(\h) - \sqrt{\frac{\calv_{\widehat\fk}(\h)}{\zeta}} -t  \le \fk(\hp) \le \cale_{\widehat\fk}(\h) + \sqrt{\frac{\calv_{\widehat\fk}(\h)}{\zeta}} + t\right) \ge 1- \zeta\right) \ge P \ge 1-\kappa,
\end{align*}
which concludes the proof.

\end{proof}

%% file: appendix/fig.tex
\begin{figure}
    \centering
    % seed = 1
    \begin{subfigure}{.23\textwidth}
        \centering
        \begin{tikzpicture}[scale=0.22/.3]
            \node (img)  {\includegraphics[width=0.9\linewidth]{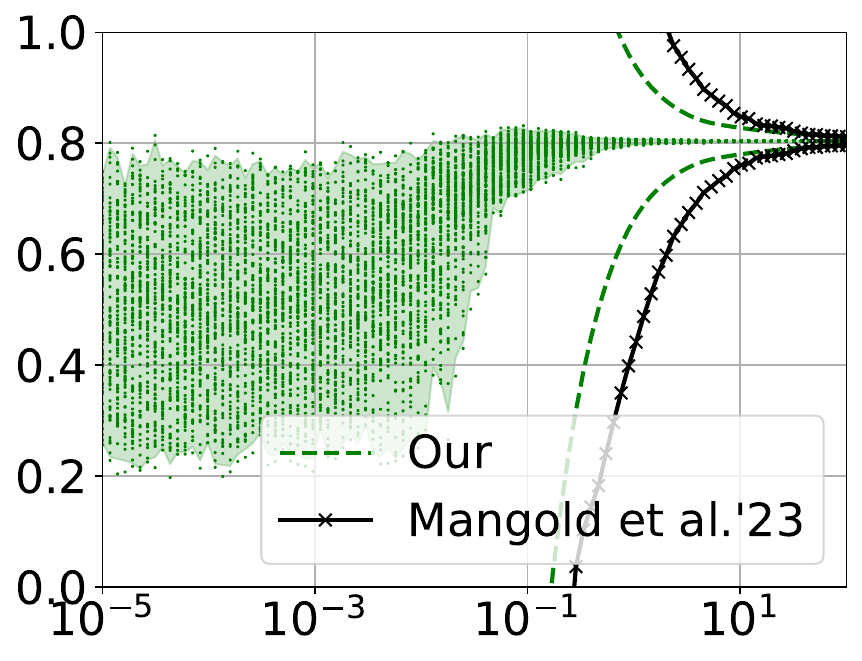}};
            \node[below=of img, node distance=0cm, yshift=3.5em,font=\color{black}] {\small $\varepsilon$};
            \node[left=of img, node distance=0cm, rotate=90, anchor=center,yshift=-3em,font=\color{black}] {\small $\acc(\hp)$};
            \node[left=of img, node distance=-1cm, rotate=90, anchor=center,yshift=-2em,font=\color{black}] {\texttt{seed = 1}};
        \end{tikzpicture}
    \end{subfigure}
    \begin{subfigure}{.23\textwidth}
        \centering
        \begin{tikzpicture}[scale=0.22/.3]
            \node (img)  {\includegraphics[width=0.95\linewidth]{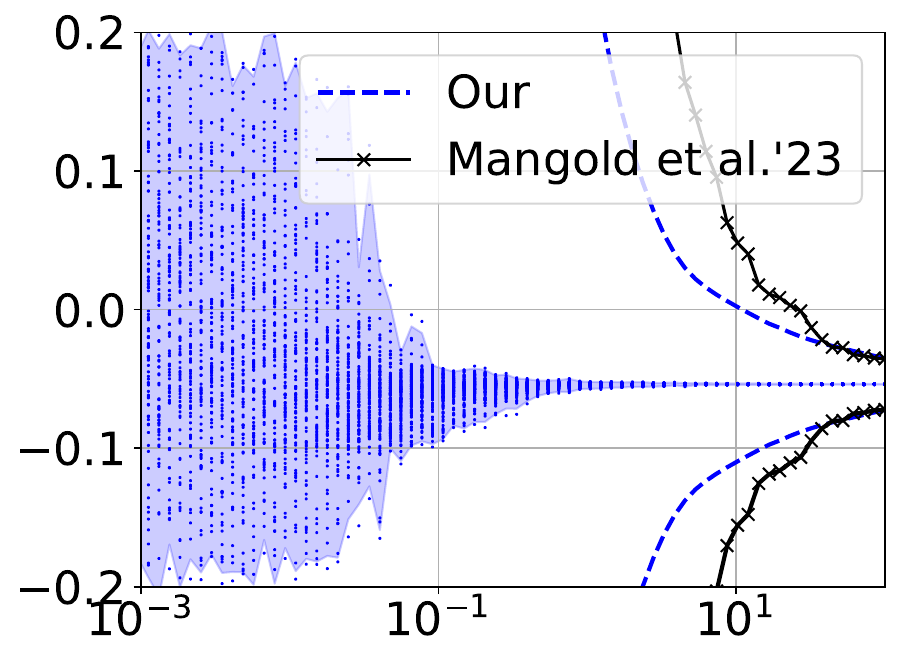}};
            \node[below=of img, node distance=0cm, yshift=3.5em,font=\color{black}] {\small$\varepsilon$};
            \node[left=of img, node distance=0cm, rotate=90, anchor=center,yshift=-3em,font=\color{black}] {\small$\calf_k(\hp)$};
            % \node at (3, 1.5) {\tiny \textcolor{blue}{$\acc(\h)$}};
        \end{tikzpicture}
    \end{subfigure}
    \begin{subfigure}{.23\textwidth}
        \centering
        \begin{tikzpicture}[scale=0.22/.3]
            \node (img)  {\includegraphics[width=0.9\linewidth]{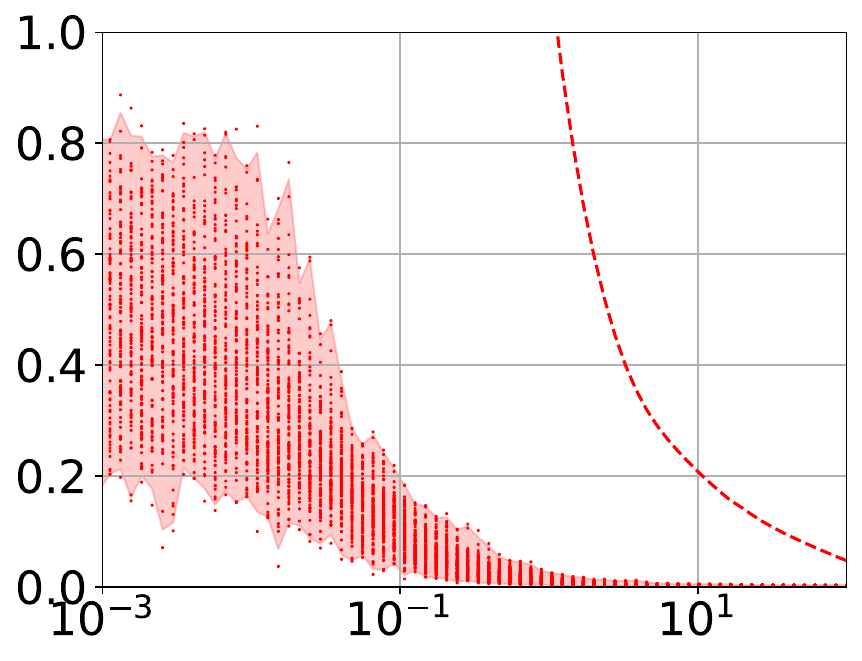}};
            \node[below=of img, node distance=0cm, yshift=3.5em,font=\color{black}] {\small $\varepsilon$};
            \node[left=of img, node distance=0cm, rotate=90, anchor=center,yshift=-3em,font=\color{black}] {\small disagreement ratio};
        \end{tikzpicture}
    \end{subfigure}
    \begin{subfigure}{.23\textwidth}
        \centering
        \begin{tikzpicture}[scale=0.22/.3]
            \node (img)  {\includegraphics[width=0.9\linewidth]{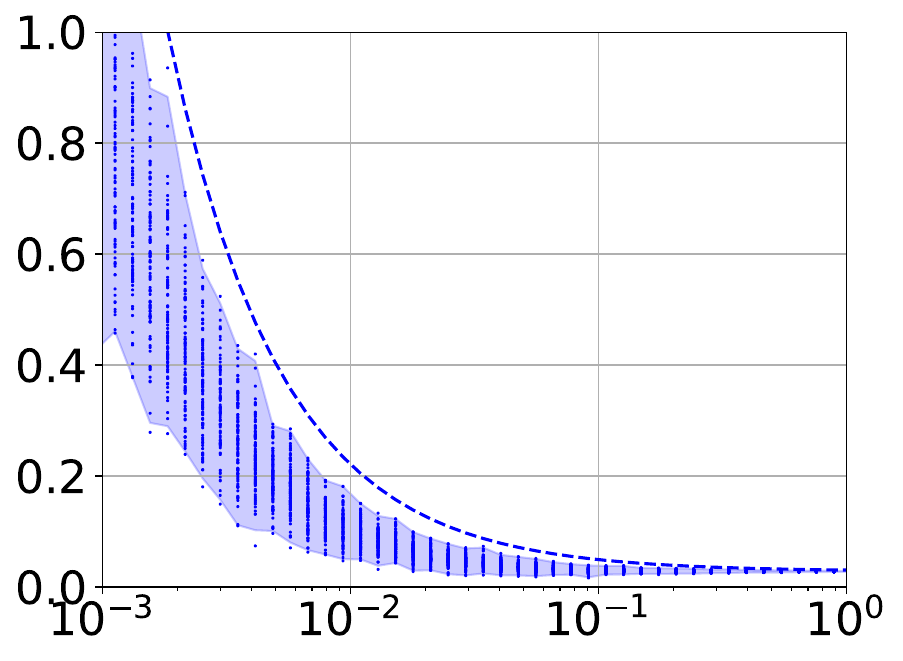}};
            \node[below=of img, node distance=0cm, yshift=3.5em,font=\color{black}] {\small$\varepsilon$};
            \node[left=of img, node distance=0cm, rotate=90, anchor=center,yshift=-3em,font=\color{black}] {\small $\|\theta^\text{priv}\|_2$};
        \end{tikzpicture}
    \end{subfigure}
% seed = 2
       \begin{subfigure}{.23\textwidth}
        \centering
        \begin{tikzpicture}[scale=0.22/.3]
            \node (img)  {\includegraphics[width=0.9\linewidth]{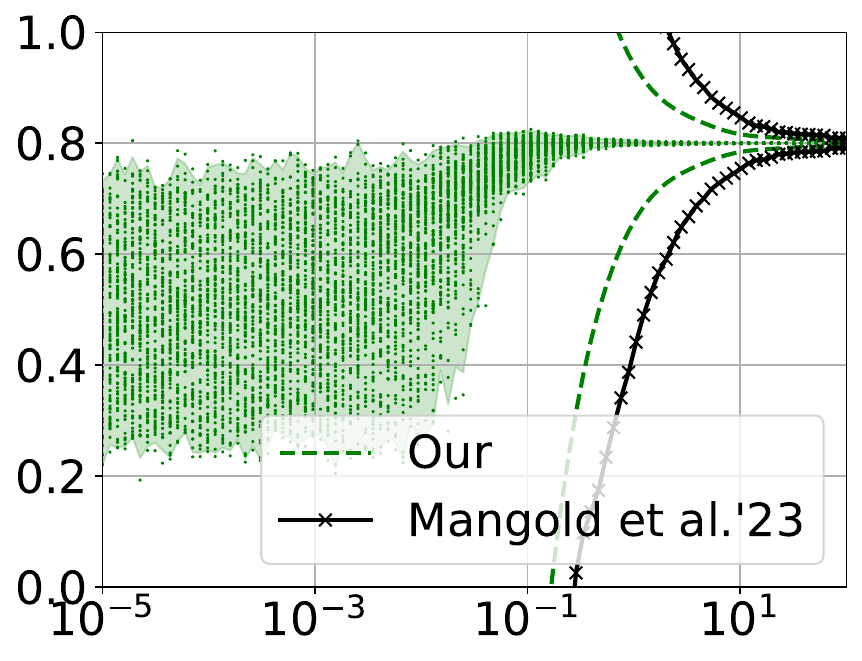}};
            \node[below=of img, node distance=0cm, yshift=3.5em,font=\color{black}] {\small $\varepsilon$};
            \node[left=of img, node distance=0cm, rotate=90, anchor=center,yshift=-3em,font=\color{black}] {\small $\acc(\hp)$};
            \node[left=of img, node distance=-1cm, rotate=90, anchor=center,yshift=-2em,font=\color{black}] {\texttt{seed = 2}};
        \end{tikzpicture}
    \end{subfigure}
    \begin{subfigure}{.23\textwidth}
        \centering
        \begin{tikzpicture}[scale=0.22/.3]
            \node (img)  {\includegraphics[width=0.95\linewidth]{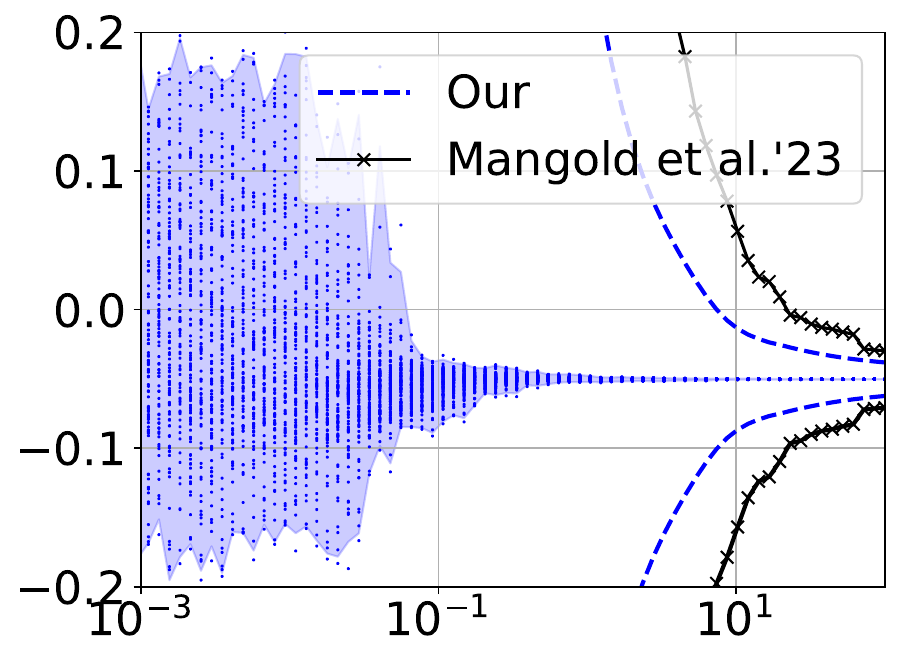}};
            \node[below=of img, node distance=0cm, yshift=3.5em,font=\color{black}] {\small$\varepsilon$};
            \node[left=of img, node distance=0cm, rotate=90, anchor=center,yshift=-3em,font=\color{black}] {\small$\calf_k(\hp)$};
            % \node at (3, 1.5) {\tiny \textcolor{blue}{$\acc(\h)$}};
        \end{tikzpicture}
    \end{subfigure}
    \begin{subfigure}{.23\textwidth}
        \centering
        \begin{tikzpicture}[scale=0.22/.3]
            \node (img)  {\includegraphics[width=0.9\linewidth]{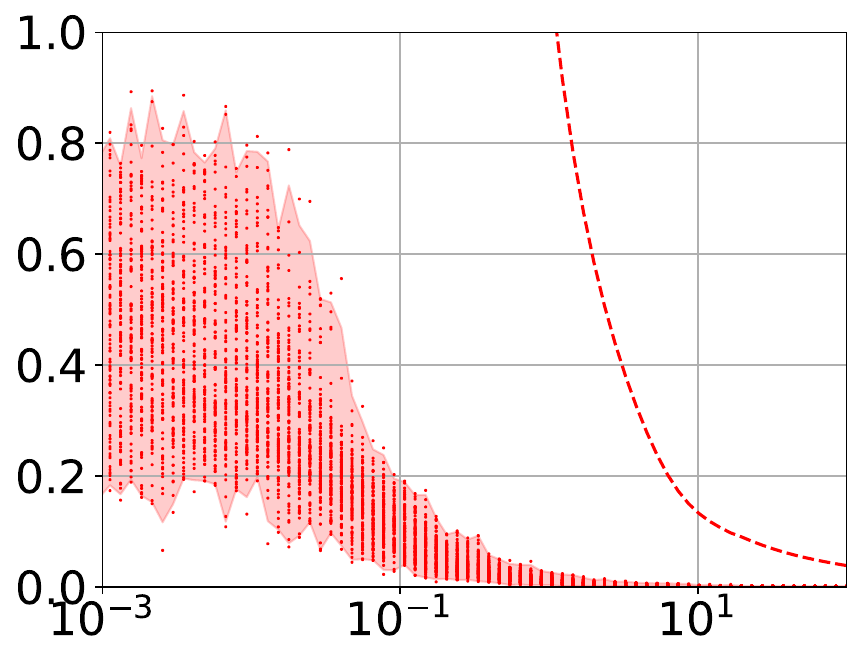}};
            \node[below=of img, node distance=0cm, yshift=3.5em,font=\color{black}] {\small $\varepsilon$};
            \node[left=of img, node distance=0cm, rotate=90, anchor=center,yshift=-3em,font=\color{black}] {\small disagreement ratio};
        \end{tikzpicture}
    \end{subfigure}
    \begin{subfigure}{.23\textwidth}
        \centering
        \begin{tikzpicture}[scale=0.22/.3]
            \node (img)  {\includegraphics[width=0.9\linewidth]{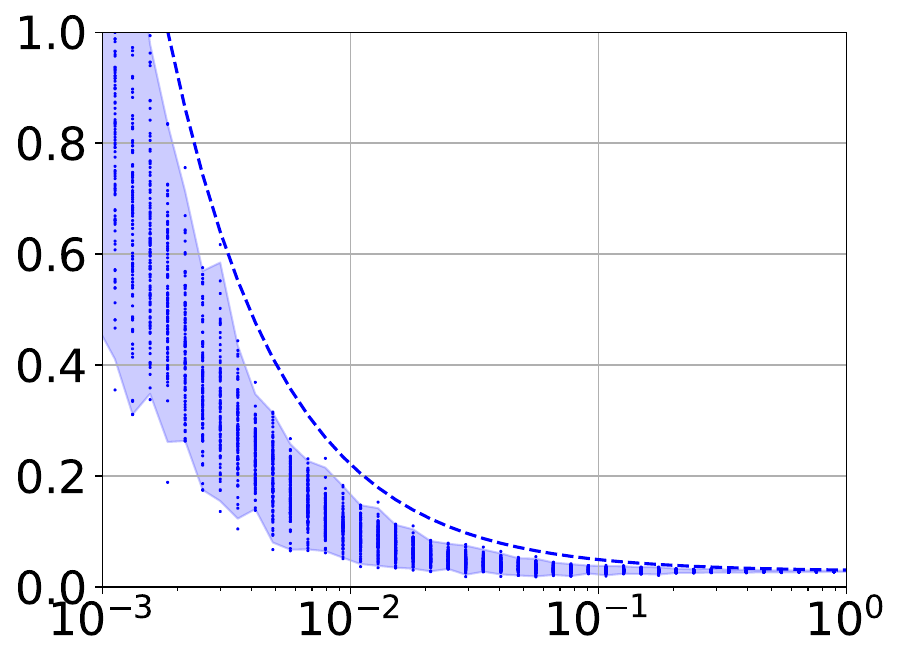}};
            \node[below=of img, node distance=0cm, yshift=3.5em,font=\color{black}] {\small$\varepsilon$};
            \node[left=of img, node distance=0cm, rotate=90, anchor=center,yshift=-3em,font=\color{black}] {\small $\|\theta^\text{priv}\|_2$};
        \end{tikzpicture}
    \end{subfigure}
    % seed = 3
           \begin{subfigure}{.23\textwidth}
        \centering
        \begin{tikzpicture}[scale=0.22/.3]
            \node (img)  {\includegraphics[width=0.9\linewidth]{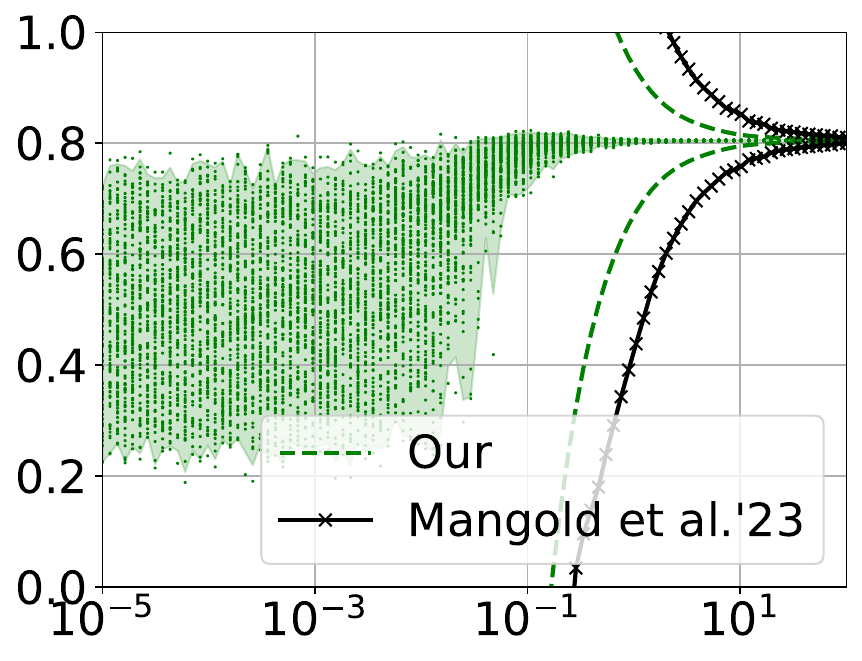}};
            \node[below=of img, node distance=0cm, yshift=3.5em,font=\color{black}] {\small $\varepsilon$};
            \node[left=of img, node distance=0cm, rotate=90, anchor=center,yshift=-3em,font=\color{black}] {\small $\acc(\hp)$};
            \node[left=of img, node distance=-1cm, rotate=90, anchor=center,yshift=-2em,font=\color{black}] {\texttt{seed = 3}};
        \end{tikzpicture}
    \end{subfigure}
    \begin{subfigure}{.23\textwidth}
        \centering
        \begin{tikzpicture}[scale=0.22/.3]
            \node (img)  {\includegraphics[width=0.95\linewidth]{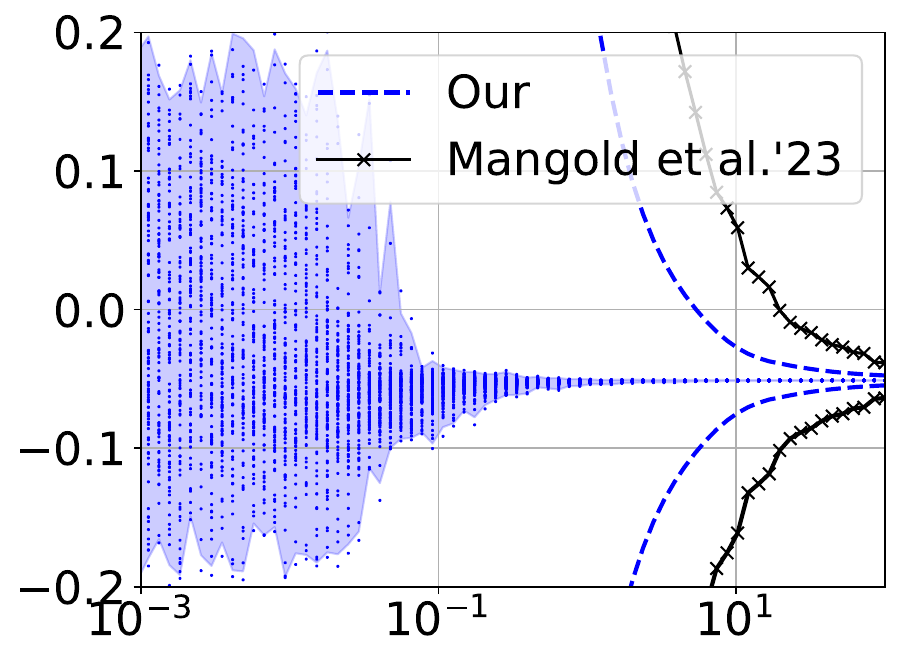}};
            \node[below=of img, node distance=0cm, yshift=3.5em,font=\color{black}] {\small$\varepsilon$};
            \node[left=of img, node distance=0cm, rotate=90, anchor=center,yshift=-3em,font=\color{black}] {\small$\calf_k(\hp)$};
            % \node at (3, 1.5) {\tiny \textcolor{blue}{$\acc(\h)$}};
        \end{tikzpicture}
    \end{subfigure}
    \begin{subfigure}{.23\textwidth}
        \centering
        \begin{tikzpicture}[scale=0.22/.3]
            \node (img)  {\includegraphics[width=0.9\linewidth]{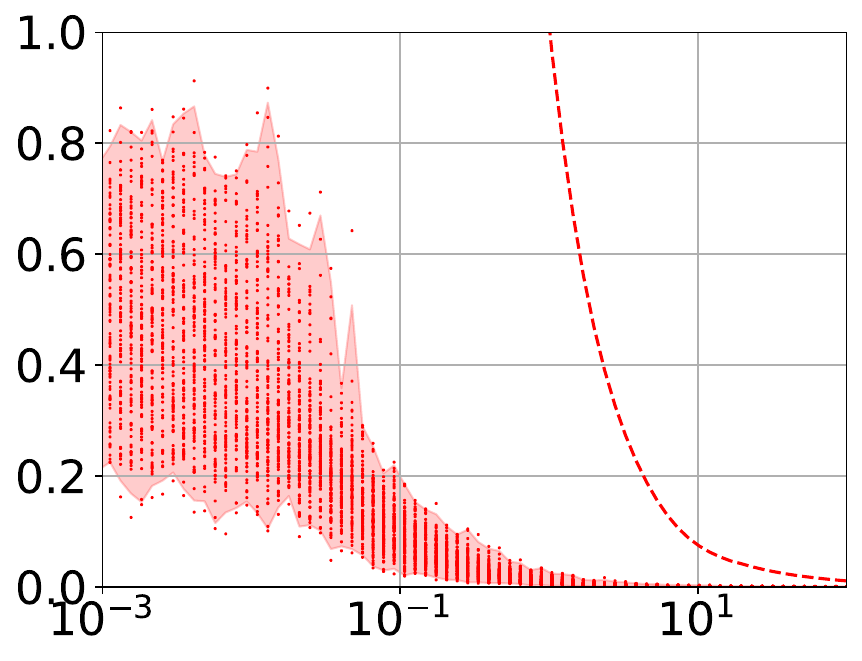}};
            \node[below=of img, node distance=0cm, yshift=3.5em,font=\color{black}] {\small $\varepsilon$};
            \node[left=of img, node distance=0cm, rotate=90, anchor=center,yshift=-3em,font=\color{black}] {\small disagreement ratio};
        \end{tikzpicture}
    \end{subfigure}
    \begin{subfigure}{.23\textwidth}
        \centering
        \begin{tikzpicture}[scale=0.22/.3]
            \node (img)  {\includegraphics[width=0.9\linewidth]{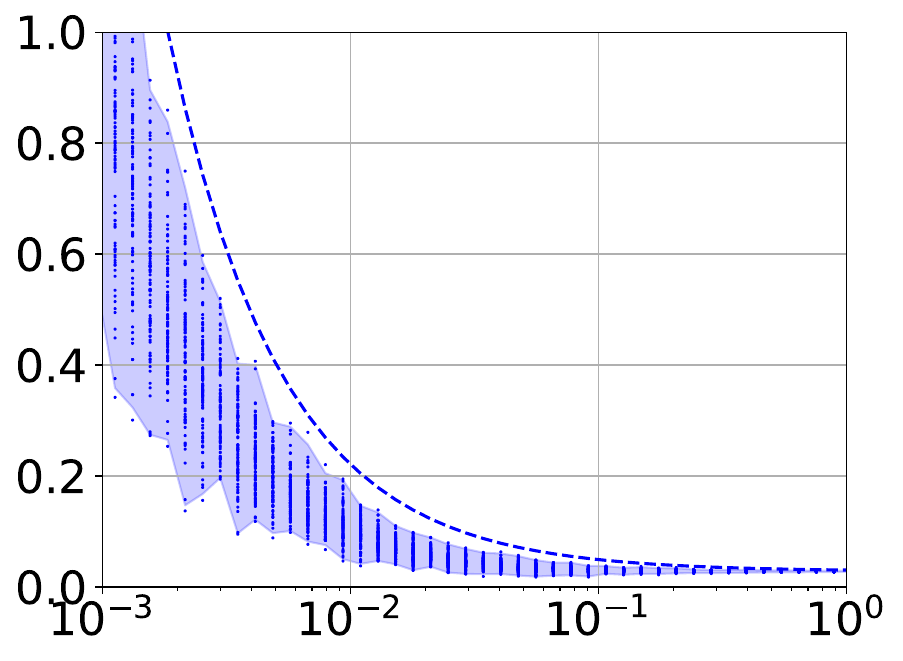}};
            \node[below=of img, node distance=0cm, yshift=3.5em,font=\color{black}] {\small$\varepsilon$};
            \node[left=of img, node distance=0cm, rotate=90, anchor=center,yshift=-3em,font=\color{black}] {\small $\|\theta^\text{priv}\|_2$};
        \end{tikzpicture}
    \end{subfigure}
    % seed = 4
           \begin{subfigure}{.23\textwidth}
        \centering
        \begin{tikzpicture}[scale=0.22/.3]
            \node (img)  {\includegraphics[width=0.9\linewidth]{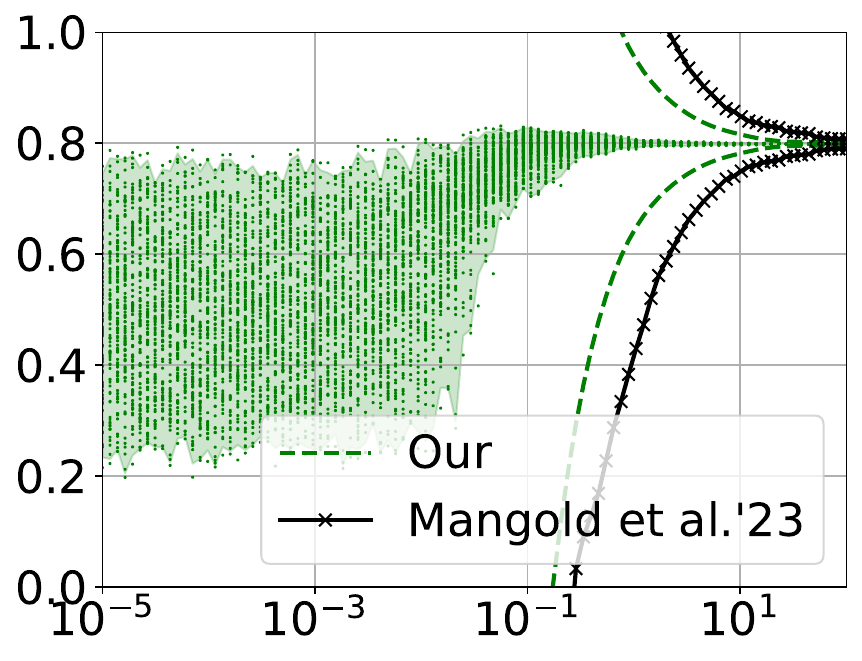}};
            \node[below=of img, node distance=0cm, yshift=3.5em,font=\color{black}] {\small $\varepsilon$};
            \node[left=of img, node distance=0cm, rotate=90, anchor=center,yshift=-3em,font=\color{black}] {\small $\acc(\hp)$};
            \node[left=of img, node distance=-1cm, rotate=90, anchor=center,yshift=-2em,font=\color{black}] {\texttt{seed = 4}};
        \end{tikzpicture}
        \caption{Accuracy}
    \end{subfigure}
    \begin{subfigure}{.23\textwidth}
        \centering
        \begin{tikzpicture}[scale=0.22/.3]
            \node (img)  {\includegraphics[width=0.95\linewidth]{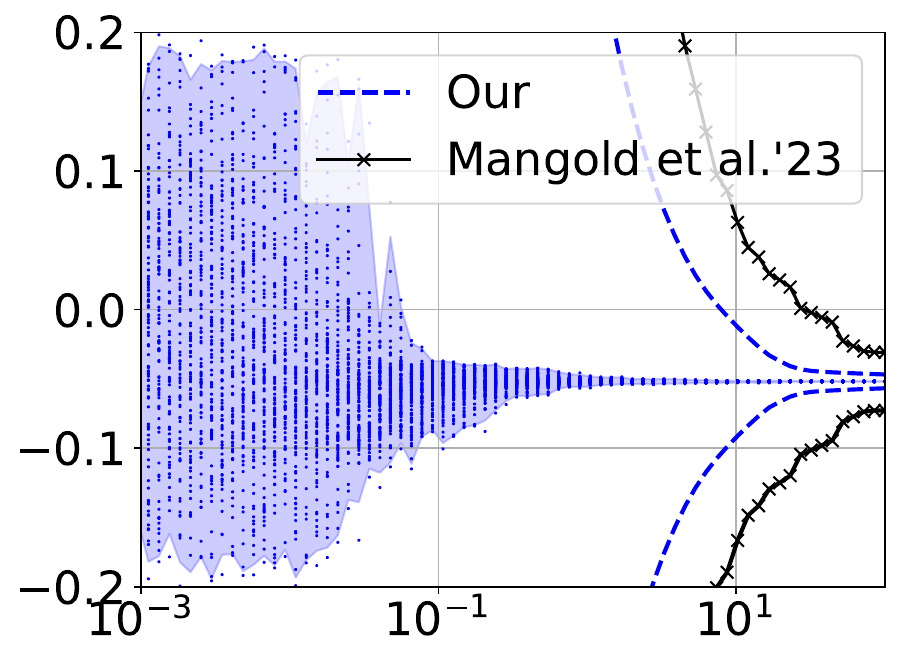}};
            \node[below=of img, node distance=0cm, yshift=3.5em,font=\color{black}] {\small$\varepsilon$};
            \node[left=of img, node distance=0cm, rotate=90, anchor=center,yshift=-3em,font=\color{black}] {\small$\calf_k(\hp)$};
            % \node at (3, 1.5) {\tiny \textcolor{blue}{$\acc(\h)$}};
        \end{tikzpicture}
        \caption{Accuracy parity}
    \end{subfigure}
    \begin{subfigure}{.23\textwidth}
        \centering
        \begin{tikzpicture}[scale=0.22/.3]
            \node (img)  {\includegraphics[width=0.9\linewidth]{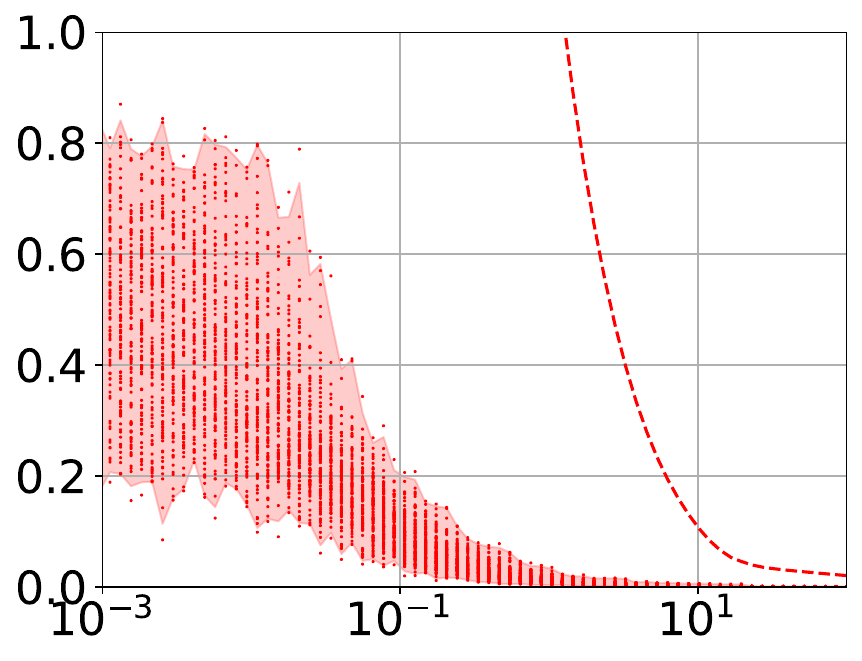}};
            \node[below=of img, node distance=0cm, yshift=3.5em,font=\color{black}] {\small $\varepsilon$};
            \node[left=of img, node distance=0cm, rotate=90, anchor=center,yshift=-3em,font=\color{black}] {\small disagreement ratio};
        \end{tikzpicture}
        \caption{Disagreement ratio}
    \end{subfigure}
    \begin{subfigure}{.23\textwidth}
        \centering
        \begin{tikzpicture}[scale=0.22/.3]
            \node (img)  {\includegraphics[width=0.9\linewidth]{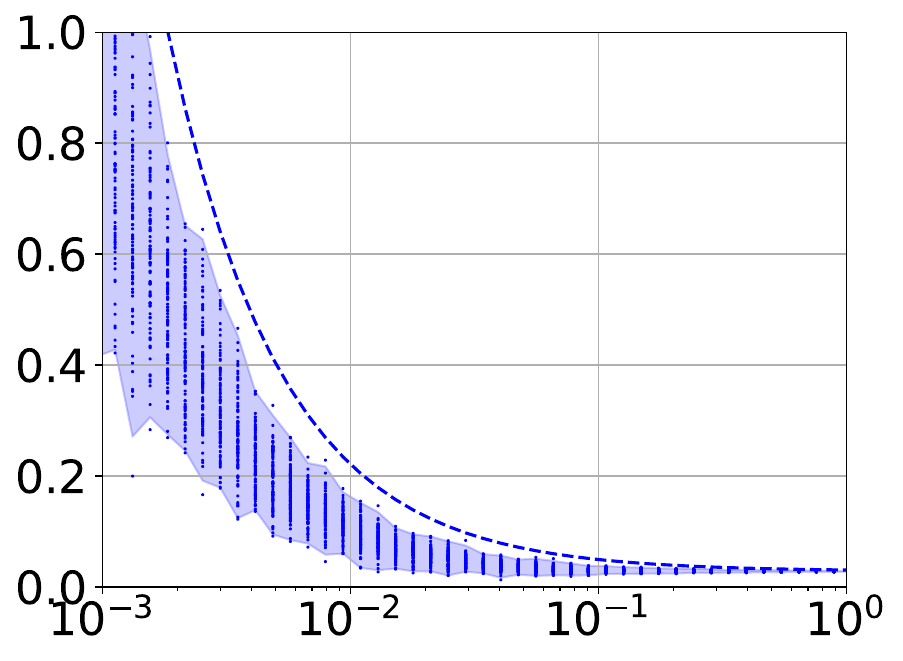}};
            \node[below=of img, node distance=0cm, yshift=3.5em,font=\color{black}] {\small$\varepsilon$};
            \node[left=of img, node distance=0cm, rotate=90, anchor=center,yshift=-3em,font=\color{black}] {\small $\|\theta^\text{priv}\|_2$};
        \end{tikzpicture}
        \caption{Individual fairness}
    \end{subfigure}
    \caption{Accuracy, accuracy parity fairness measure, disagreement ratio and individual fairness of private models  $\hp$ for different values of $\varepsilon$ on \texttt{Adult} dataset.  Different rows correspond to different values of random seeds ($1$, $2$, $3$, $4$). The $99\%$-confidence bounds are shown by dashed and crossed lines, and color-filled regions correspond to regions where $99\%$ of measurements lie.}
    \label{fig: bounds extra}
\end{figure}
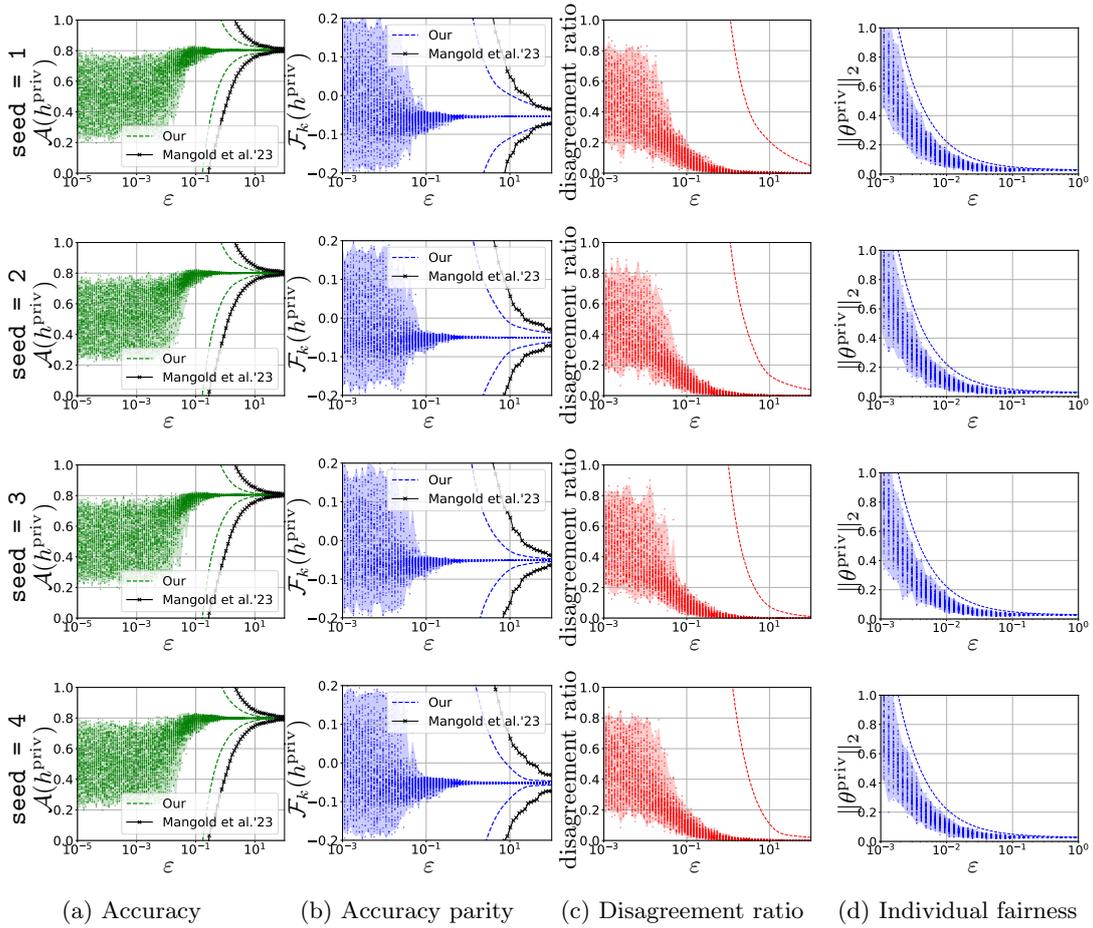

In this section, we provide additional experimental results. In Fig.~\ref{fig: bounds extra} we illustrate the high-probability bounds from Section~\ref{sec:individual fairness}, Section~\ref{sec:disagreement}, and Section~\ref{sec:bound} for different values of random seeds.
